\documentclass{article} 
\usepackage{iclr2027_conference,times}
\iclrfinalcopy

\usepackage{amsmath,amsfonts,bm}

\def\eqref#1{equation~\ref{#1}}

\def\1{\bm{1}}

\DeclareMathAlphabet{\mathsfit}{\encodingdefault}{\sfdefault}{m}{sl}
\SetMathAlphabet{\mathsfit}{bold}{\encodingdefault}{\sfdefault}{bx}{n}

\usepackage{hyperref}       
\usepackage{url}            
\usepackage{booktabs}       
\usepackage{amsfonts}       
\usepackage{nicefrac}       
\usepackage{microtype}      
\usepackage{xcolor}         
\usepackage{algorithm}       
\usepackage{algpseudocode}   
\usepackage{amsmath}
\usepackage{amssymb}
\usepackage{mathtools}
\usepackage{amsthm}

\usepackage{hyperref}
\usepackage{url}
\usepackage{booktabs}
\usepackage{graphicx}
\usepackage{subcaption}
\usepackage{multicol}
\usepackage{makecell}
\usepackage{multirow}
\usepackage{cancel}
\usepackage{tcolorbox}
\usepackage{amsmath} 
\usepackage{siunitx}
\usepackage{cleveref}
\usepackage{booktabs}
\usepackage{multirow}
\usepackage{colortbl}
\usepackage{collcell}
\usepackage{arydshln}
\usepackage{tabularx}
\usepackage{lipsum}
\usepackage{wrapfig}
\usepackage{pifont}
\usepackage{capt-of}
\usepackage{bm}
\usepackage{lscape}
\usepackage{tabularx}
\usepackage{pifont}
\usepackage{xcolor}
\usepackage{amssymb}
\usepackage{float}
\usepackage{wrapfig}
\usepackage{amsthm}

\newtheorem{theorem}{Theorem}
\definecolor{darkgreen}{rgb}{0.0, 0.5, 0.0} 
\definecolor{darkred}{rgb}{0.5, 0.0, 0.0}

\title{Direct Trajectory Optimization for Inference-Time Alignment of Text-to-Image Diffusion Models}

\author{Huixuan Zhang, Li Lin, Junzhe Zhang \& Xiaojun Wan  \\
Wangxuan Institute of Computer Technology, Peking University \\
\texttt{\{zhanghuixuan, efsotr\_l, junzhezhang\}@stu.pku.edu.cn, wanxiaojun@pku.edu.cn} \\
}

\begin{document}

\maketitle

\begin{abstract}
Alignment of text-to-image diffusion models with reward functions is a key challenge in controllable generation. Existing approaches typically rely on optimizing model inputs or fine-tuning denoising networks, which requires computationally expensive backpropagation through large denoising models. In this work, we propose Direct Trajectory Optimization (DTO), a training-free framework that performs alignment by directly optimizing the sampling trajectory of diffusion models. Specifically, DTO treats intermediate denoising scores and sampled noise as the optimization variables, eliminating the need for gradients through the denoising network. To mitigate the out-of-distribution problem, we introduce Reward-Optimized Score Guidance, which interpolates between the optimized scores and the pretrained model predictions to stabilize sampling, while providing a provable theoretical guarantee of reward improvement under constrained distributions. Empirically, DTO improves alignment performance across three diffusion models on two benchmarks, outperforming existing inference-time alignment methods while maintaining high efficiency. These results suggest that trajectory-level optimization is a promising direction for efficient and effective diffusion model alignment.
\end{abstract}
\section{Introduction}
Text-to-image diffusion models have achieved remarkable success in generating high-quality images aligned well with textual prompts \citep{esser2024scaling, wu2025qwenimagetechnicalreport, seedream2025seedream40nextgenerationmultimodal}. However, despite their strong generative capabilities, these models often struggle to fully satisfy fine-grained user preferences or task-specific objectives, such as improved overall image quality or stronger alignment with textual prompts. This motivates the problem of alignment: steering a pretrained diffusion model toward generating samples that maximize a given reward function.

A common approach to this problem is to fine-tune the denoising network using reinforcement learning or preference optimization techniques \citep{gu2024diffusionrpoaligningdiffusionmodels, ddpo, wallace2023diffusiondpo}. While effective in improving alignment, these methods typically require substantial computational resources and retraining for different reward functions, which limits their flexibility in practical scenarios. To avoid retraining, recent training-free approaches perform optimization at inference time by modifying model inputs such as noise, latent variables, or textual embeddings \citep{tang2024inference, na2025diffusion}. Although these methods eliminate parameter updates, they still rely on backpropagation through the denoising network, which introduces considerable computational overhead and can complicate the optimization process.

To reduce computational cost and improve alignment performance, we propose Direct Trajectory Optimization (DTO). \textbf{DTO directly optimizes the sampling trajectory (including the sampled noise and denoising scores at each step), eliminating the need for gradients through the denoising model.} This perspective decouples alignment optimization from the gradients of internal parameters of the denoising network, enabling efficient and effective reward-driven refinement at inference time.

However, directly optimizing the trajectory introduces the out-of-distribution problem, as unconstrained updates may disrupt the pretrained diffusion distribution and lead to degraded quality. To address this, we design Reward-Optimized Score Guidance (ROSG) in DTO, which interpolates between the optimized scores and the pretrained model's score predictions. We further theoretically show that the trajectory obtained via ROSG preserves reward improvement under constrained distributions, achieving a balance between distribution maintenance and reward optimization.

\begin{figure}
    \centering
    \begin{subfigure}[b]{0.23\textwidth}
        \centering
        \includegraphics[width=1.0\linewidth]{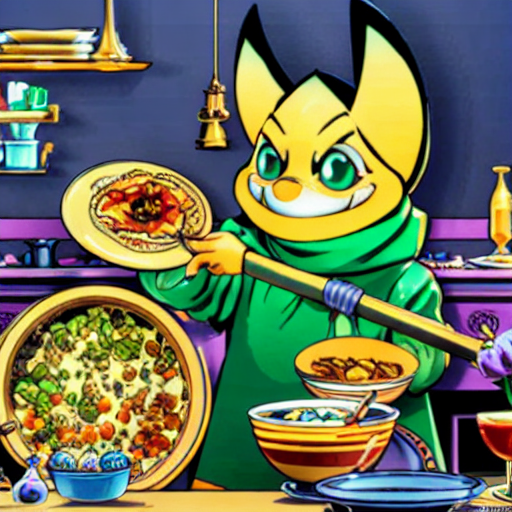}
        \caption{SD1.5+ReNeg \citep{li2025reneg}}
    \end{subfigure}
    \hfill
    \begin{subfigure}[b]{0.23\textwidth}
        \centering
        \includegraphics[width=1.0\linewidth]{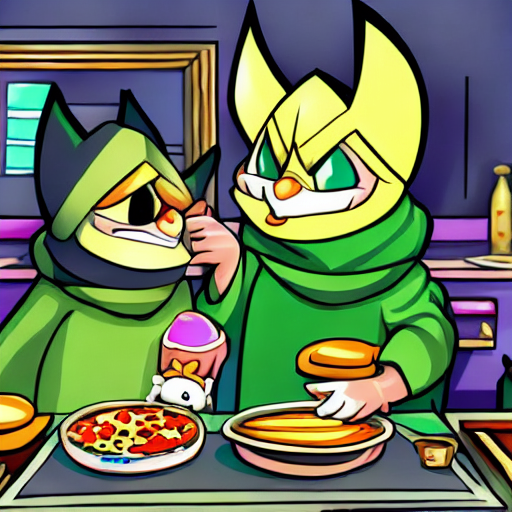}
        \caption{SD1.5+DATE \citep{na2025diffusion}}
    \end{subfigure}
    \hfill
    \begin{subfigure}[b]{0.23\textwidth}
        \centering
        \includegraphics[width=1.0\linewidth]{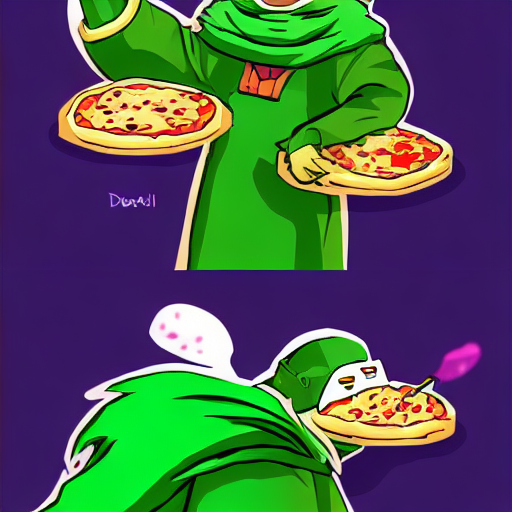}
        \caption{SD1.5+DNO \citep{tang2024inference}}
    \end{subfigure}
    \hfill
    \begin{subfigure}[b]{0.23\textwidth}
        \centering
        \includegraphics[width=1.0\linewidth]{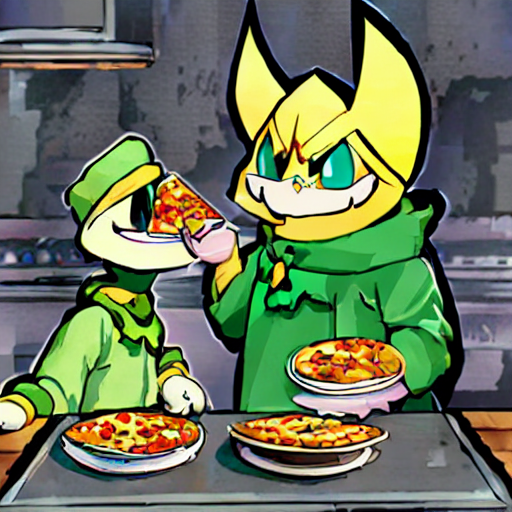}
        \caption{SD1.5+DTO(Ours)}
    \end{subfigure}
    \caption{Generation results of different alignment methods. The prompt is ``Ralsei and Asriel from Deltarune eating pizza.''}
     \label{fig:intro_case}
\end{figure}

Experimental results demonstrate that DTO improves alignment performance across three diffusion models on two benchmarks, outperforming existing inference-time alignment methods while maintaining high computational efficiency. We further show that DTO achieves a better time-performance trade-off compared with other inference-time methods. Ablation studies demonstrate that both trajectory optimization and Reward-Optimized Score Guidance improve performance, confirming the effectiveness of each component. Human evaluation further suggests that DTO outperforms other methods. A qualitative case comparison is provided in Figure \ref{fig:intro_case}.

Our contributions are summarized as follows:
\begin{enumerate}
\item We propose Direct Trajectory Optimization (DTO), a novel framework that directly optimizes the diffusion sampling trajectory, including sampled noise and denoising score, without back-propagation through the denoising network.
\item We introduce Reward-Optimized Score Guidance to mitigate the out-of-distribution problem, and provide theoretical analysis showing reward improvement under constrained distribution.
\item Experiments demonstrate that DTO achieves a better trade-off between optimization performance and computational efficiency than existing methods across diverse settings, achieving 10$\times$ higher speed than previous SoTA alignment method DNO while still providing better alignment performance.
\end{enumerate}

\section{Related Works}

\subsection{Text-to-Image Diffusion Models}
DDPM was first introduced in \cite{NEURIPS2020_DDPM}, serving as the foundation for subsequent diffusion models. For incorporating textual condition into diffusion models, there are two most commonly used guidance strategy: classifier guidance \citep{dhariwal2021diffusion} and classifier-free guidance \citep{ho2022classifier}, while the latter one is more widely used nowadays. Denoising in latent space was proposed by \cite{Rombach_2022_CVPR}, proven highly successful. Different noise scheduling has been introduced in diffusion models \citep{nichol2021improved, song2021DDIM, song2021scorebased, karras2022elucidating, kingma2024understanding, lu2022dpm, lu2022dpm++, albergo2022building, liu2022flow, lipman2022flow, esser2024scaling}. 

Based on these studies, many practical text-to-image diffusion models have been introduced, including \citep{Rombach_2022_CVPR, podell2023sdxl, esser2024scaling, wu2025qwenimagetechnicalreport, chen2024pixartsigma, flux2024}. There are also powerful closed-source text-to-image models, including \citep{seedream2025seedream40nextgenerationmultimodal, gao2025seedream}. However, since inference details of these models are not publicly available, we focus on open-source diffusion models for fair comparison.

\subsection{Alignment Problem of Text-to-Image Diffusion Models}
Applying a reward model for aligning and improving diffusion models is a popular topic. 
One approach is training-based methods, including DDPO \citep{ddpo}, which is inspired by RLHF \citep{christiano2017deep, ouyang2022training}, and Diffusion-DPO \citep{wallace2023diffusiondpo}, following direct preference optimization (DPO) \citep{rafailov2023direct}. Regarding differentiable reward models, directly finetuning diffusion models is also a promising approach as introduced by ReFL \citep{xu2023imagereward}. However, these methods all require tuning model parameters, resulting in high computational cost and low flexibility.

There are also training-free alignment strategies, including direct noise optimization (DNO) \citep{tang2024inference}, diffusion adaptive text embedding (DATE) \citep{na2025diffusion}, and ReNeg \citep{li2025reneg}. Despite their different design choices, their key insight all lies in using a differentiable reward model for optimizing part of the input during inference. Though eliminating the need for tuning model parameters, these methods still require back-propagation through the denoising model, leading to difficulty in optimization and still potentially high cost. 

Regarding reward models, classical reward models include ImageReward \citep{xu2023imagereward}, HPS \citep{wu2023hps}, and PickScore \citep{kirstain2023pickscore}. Recent reward models are built on large MLLMs, including UnifiedReward \citep{unifiedreward, unifiedreward-think, unifiedreward-flex}, and Minos \citep{zhang2025minosmultimodalevaluationmodel}.

\section{Preliminaries: Diffusion Models}
\label{sec:bg}
Starting from an initial noise $x_T$, diffusion model iteratively refines the sample towards a clean output $x_0$. During inference, at each timestep $t$, a denoising network $\epsilon_\theta(x_t, t)$ predicts a score used to guide the generation. For conditional generation with a given condition $c$, classifier-free guidance (CFG) \citep{ho2022classifier} combines conditional and unconditional predictions, and following common practice in negative prompt guidance, CFG can be naturally extended to incorporate negative conditions $n$ :
\begin{equation}
\tilde{\epsilon}_{\theta}(x_t, t, c, n) = \epsilon_{\theta}(x_t, t, n) + w \left[ \epsilon_\theta(x_t, t, c) - \epsilon_\theta(x_t, t, n) \right]
\end{equation}

Noise scheduling is formulated following SDE or ODE \citep{karras2022elucidating}. For an ODE-based noise scheduler $F$, where $F$ denotes a single-step numerical solver, let $\epsilon_{t} = \tilde{\epsilon}_{\theta}(x_{t}, t, c, n)$, the update follows:
\begin{equation}
x_{t-1} = F\left(x_t, \epsilon_{t}, t\right)
\label{eq:noise_schedule}
\end{equation}

An SDE-based noise scheduler additionally involve sampling a random noise $z_{t}$ at each timestep ($x_{t-1} = F\left(x_t, \epsilon_{t}, t, z_{t}\right)$). By iteratively applying Equation \ref{eq:noise_schedule}, the denoised result $x_{0}$ could be achieved. For latent diffusion models, a decoder $D$ decodes this $x_0$ to real images $D(x_{0})$. \footnote{Since $D$ is differentiable and always fixed, we omit $D$ in the following discussion and directly use $x_0$ to represent the final generated image. All rewards are computed on decoded images unless otherwise specified.} We denote the entire framework (the noise scheduler $F$, the denoising network $\epsilon_{\theta}$, and the decoder $D$) as the overall diffusion model $M$. $M$ takes the initial noise sampled $x_{T}$, a positive condition $c$, a (possible) negative condition $n$, and a set of sampled noise $\{z_{t}\}_{t=1}^{T}$ (for SDE-based noise schedulers) as input, and produces a generated image as an output. 

\section{Direct Trajectory Optimization}

\subsection{Problem Definition and Motivation}

\begin{table}[htbp]
\centering
\caption{Comparison of optimization formulations of existing methods.}
\begin{tabular}{c|c}
\toprule
Method & Optimization Formulation \\
\midrule
DNO \citep{tang2024inference} & $\max_{x_T, \{z_{t}\}} R(M(x_T, c, n, \{z_t\}), c)$ \\
DATE \citep{na2025diffusion} & $\max_{\tilde{c}} R(M(x_T, \tilde{c}, n, \{z_t\}), c)$ \\
ReNeg \citep{li2025reneg} & $\max_{n} R(M(x_T, c, n, \{z_t\}), c)$ \\
\bottomrule
\end{tabular}
\label{tab:method_comp}
\end{table}

Given a pretrained diffusion model $M$ and a reward function $R$, test-time alignment aims to steer the generation process toward samples that maximize the reward under a given text condition $c$:

\begin{equation}
\max R\left(M(x_T, c, n, \{z_t\}_{t=1}^{T}), c\right).
\end{equation}

Existing methods mainly differ in the choice of \textbf{optimization variables within the diffusion pipeline}, as summarized in Table~\ref{tab:method_comp}. Despite these differences, a common characteristic of these methods is that they rely on gradients propagated through the denoising network. Let $u$ denote a generic set of optimized variables; most approaches can be expressed in the form of gradient-ascent updates:
\begin{equation}
u' = u + \eta \frac{\partial R(M(x_T, \tilde{c}, n, \{z_t\}), c)}{\partial u}.
\label{eq:ga}
\end{equation}

However, such requirement introduces two key challenges. First, this may restrict the effective optimization space and complicate the optimization process. Second, it leads to considerable computational overhead, especially for large-scale diffusion models. A detailed discussion of the gradient flow of previous methods can be found in Appendix \ref{sec:app-gf}.

These limitations motivate a natural question: \textbf{Can we perform test-time alignment without back-propagating through the denoising network?}

\subsection{Optimizing Trajectory}

In the following discussion, we focus on ODE-based noise schedulers to present our method, as they are widely used in modern diffusion systems \citep{esser2024scaling, wu2025qwenimagetechnicalreport}. Our formulation naturally extends to SDE-based samplers, which are discussed in Appendix \ref{sec:app-sde} and \ref{sec:app-sde-exp}.

As shown in Equation \ref{eq:noise_schedule}, given a noise scheduler $F$, the diffusion generation process is fully determined by the initial noise $x_T$ and the sequence of intermediate predicted scores $\{\epsilon_t\}_{t=1}^{T}$. Existing methods implicitly influence these variables by optimizing inputs to the denoising network $\epsilon_\theta$, thereby indirectly shaping the induced trajectory of the sampling process.

\begin{algorithm}[htbp]

\caption{Direct Trajectory Optimization (DTO)}
\label{algo:dto}
\begin{algorithmic}[1]
\Require
Denoising model $\epsilon_{\theta}$, Noise Scheduler $F$ (corresponding to $f$), total denoising steps $T$, Reward model $R$, learning rate $\eta$, optimization steps $S$, weight $w_{t}$, text condition $c$. 

\State Sample an initial noise $x_T$ and the initialize the trajectory $\tau=(x_{T}, \epsilon_{T}, ...,\epsilon_{1})$ using $\epsilon_{\theta}, F, x_{T}, c$
\State Calculate the current reward $r = R(f(\tau))$, cache best trajectory $\tau^{*}\leftarrow \tau$.
\For{$step=1$ to $S$}
    \State $\tau \leftarrow \tau + \eta \nabla_{\tau}R(f(\tau))$ \Comment{Gradient ascent.}
    \Statex $\triangleright$ \textit{Reward-Optimized Score Guidance.}
    \For{$t=T$ to $1$}
        \State $\epsilon_{t}' = \tilde{\epsilon}_{\theta}(x_{t}, t,c,n)$ 
        \State $\hat{\epsilon}_{t}\leftarrow\epsilon'_{t}+w_{t}(\epsilon_{t}-\epsilon'_{t})$
        \State $x_{t-1}=F(x_{t}, \hat{\epsilon}_{t}, t)$
    \EndFor
    \State $\tau \leftarrow (x_{T}, \hat{\epsilon}_{T},...,\hat{\epsilon}_{1})$
    \Statex $\triangleright$ \textit{Select the best trajectory.}
    \State Calculate current reward $r_{c} = R(f(\tau))$
    \If{$r_{c}>r$}
        \State $r \leftarrow r_{c}, \tau^{*} \leftarrow \tau$ 
    \EndIf
\EndFor
\State \Return Optimized sample $x_{0}^{*}=f(\tau^{*})$
\end{algorithmic}
\end{algorithm}

In contrast, we take a different perspective: \textbf{instead of treating $\{\epsilon_t\}$ as outputs of a fixed network, we view them as explicit variables that define the diffusion trajectory}. We actually relax the functional dependency induced by the pretrained denoising network and treat these intermediate scores as free variables able to be optimized, whose gradients flow only through the noise scheduler. From this viewpoint, the denoising network $\epsilon_\theta$ can be interpreted as a mechanism that provides an initialization of these variables, rather than a constraint on their optimization. This motivates our proposed method, \textbf{Direct Trajectory Optimization (DTO)}, whose overview is provided in Algorithm \ref{algo:dto}.

Formally, we define a diffusion trajectory as:
\begin{equation}
\tau = (x_T, \epsilon_T, \ldots, \epsilon_1),
\end{equation}
where $x_T$ and $\{\epsilon_t\}_{t=1}^{T}$ parameterize the entire sampling process induced by a fixed noise scheduler.

Given a fixed noise scheduler $F$, the final output can be expressed as a deterministic function $f$ (by unrolling the iteration of Equation~\ref{eq:noise_schedule}) of the trajectory:
\begin{equation}
x_0 = f(\tau),
\end{equation}

Based on this formulation, for simplicity let $R(f(\tau)) = R(f(\tau), c) = R(f(x_T, \epsilon_T, \ldots, \epsilon_1), c)$, the optimization target can be written as:
\begin{equation}
\max_{\tau} \; R(f(\tau))
\label{eq:dto}
\end{equation}

Since $f$ is differentiable with respect to its inputs, gradients can be propagated through the noise scheduler:
\begin{align}
\frac{\partial R}{\partial \epsilon_t} &= \frac{\partial R}{\partial x_{t-1}} \frac{\partial F(x_t, \epsilon_t, t)}{\partial \epsilon_t}, \\
\frac{\partial R}{\partial x_t} &= \frac{\partial R}{\partial x_{t-1}} \frac{\partial F(x_t, \epsilon_t, t)}{\partial x_t}, \\
\nabla_{\tau} R(f(\tau)) &=\left(
\frac{\partial R}{\partial x_T},
\frac{\partial R}{\partial \epsilon_T},
\ldots,
\frac{\partial R}{\partial \epsilon_1}
\right).
\label{eq:dto_grad}
\end{align}

Given this gradient, standard gradient ascent (Equation \ref{eq:ga}) can be naturally applied as the optimization strategy. Since this optimization can be conducted multiple steps, we can select the trajectory $\tau^{*}$
  that achieves the highest reward during the process.

From this perspective, the trajectory $\tau$ can actually be initialized arbitrarily. However, a poorly initialized trajectory is difficult to optimize and may lead to poor convergence. Therefore, in practice, we initialize 
$\tau$ using the standard diffusion process described in Section \ref{sec:bg}, which provides a trajectory that is consistent with the pretrained model distribution and serves as a strong starting point for optimization. Starting from this initialization, we perform direct optimization over the trajectory variables, freeing the scores $\epsilon_t$  from the constraints of the denoising network. This formulation removes the dependence on gradients through the denoising network, $\frac{\partial \tilde{\epsilon}_\theta}{\partial x_t}$, and instead operates entirely through the noise scheduler dynamics. This significantly reduces the computational overhead of backpropagation through large neural networks. We provide a detailed discussion of gradient flow in Appendix \ref{sec:app-gf}.

\subsection{Out-of-Distribution (OOD) Problem}

While naive direct trajectory optimization in Equation~\ref{eq:dto} improves efficiency and flexibility, it introduces a critical challenge: the out-of-distribution (OOD) problem, a common issue in inference-time optimization methods \citep{tang2024inference}. Since the optimized trajectory is no longer fully constrained by the pretrained diffusion model, it may drift away from the pretrained distribution, leading to reward hacking and degraded generation quality.

Motivated by CFG \citep{ho2022classifier} and adaptive guidance \citep{zhang2025much}, we mitigate this issue by pulling the optimized score $\epsilon_t$  back toward the pretrained model distribution after each optimization step, which serves as a form of guidance. We refer to this strategy as \textit{Reward-Optimized Score Guidance} (ROSG). Formally, for each timestep $t$, we compute:
\begin{align}
\epsilon'_t &= \tilde{\epsilon}_\theta(x_t, t, c, n), \\
\hat{\epsilon}_t &= \epsilon'_t + w_t (\epsilon_t - \epsilon'_t), \\
x_{t-1} &= F(x_t, \hat{\epsilon}_t, t).
\label{eq:adaptive}
\end{align}

This formulation can be interpreted as a convex interpolation between the pretrained model prediction and the optimized score, thereby balancing reward maximization and distribution preservation. Following a similar motivation to that in \citep{zhang2025much}, we define $w_t$ as
\begin{equation}
w_{t} =
\begin{cases}
1, & t < \frac{1}{2}T \\
1 - \frac{t}{T}, & t \geq \frac{1}{2}T.
\end{cases}
\label{eq:scale_weight}
\end{equation}

Comparisons with alternative $w_t$ schedules are given in Appendix~\ref{sec:app-abla}.

\paragraph{Theoretical Guarantee}

We further provide a theoretical justification that such ROSG does not significantly hinder optimization. Specifically, we have:

\begin{theorem}
Let $\hat{\tau}$ denote the trajectory achieved using gradient ascent and Reward-Optimized Score Guidance, $\tau$ denote trajectory before ROSG and optimization. Assume the reward function $R(f(\tau))$ is $L$-Lipschitz continuous. Then:
\begin{equation}
R(f(\hat{\tau})) \geq R(f(\tau)) + \eta ||\nabla_{\tau} R(f(\tau))||^{2}-
L \sum_{t=1}^{T} (1 - w_t)||\epsilon'_t - \epsilon_t|| + O(\eta^{2}).
\end{equation}
\label{theo:combine}
\end{theorem}

Theorem~\ref{theo:combine} implies that when the optimized score $\epsilon_t$  remains sufficiently close to the pretrained prediction $\epsilon_t'$, and the reward gradient magnitude is large, the trajectory achieved using ROSG $\hat{\tau}$  still yields an increase in reward. Conversely, when the gradient norm is small, the optimization signal is weak, indicating limited potential for further improvement. Meanwhile, when the deviation between $\epsilon_t$  and $\epsilon_t'$  becomes large, the trajectory is likely to suffer from severe OOD issues, in which case aggressive optimization may become counterproductive. The proof together with a more detailed discussion is given in Appendix \ref{sec:app-proof}. Score-drift and FID measurements that probe this regime are reported in Appendix~\ref{sec:app-abla}, and a corresponding qualitative failure without ROSG is shown in Appendix~\ref{sec:app-case}.

\section{Experiments and Analysis}
\subsection{Experiment Setup}

\paragraph{Datasets}
We evaluate our method on two widely-used public benchmarks: the HPD-v2 test set (400 prompts) \citep{wu2023hps} and GenAIBench (1,600 prompts) \citep{li2024evaluating}.

\paragraph{Baseline Methods}
We evaluate our method on three open-source diffusion models, including Stable-Diffusion-1.5 (SD1.5) \citep{Rombach_2022_CVPR}, Stable-Diffusion-XL (SDXL) \citep{podell2023sdxl}, and Stable-Diffusion-3 (SD3) \citep{esser2024scaling}. Results on Stable-Diffusion-3.5-Large-Turbo and FLUX.1-dev are reported in Appendices~\ref{sec:app-backbones} and \ref{sec:app-latency}. We compare against several representative test-time alignment methods, including DNO \citep{tang2024inference}, ReNeg \citep{li2025reneg}, and DATE \citep{na2025diffusion}. Among them, ReNeg and DATE are implemented on SD1.5 only, while DNO is implemented on both SD1.5 and SDXL, following their official implementations. In addition, we include Best-of-N sampling (BoN) as a strong  baseline.

\paragraph{Implementation Details}
Following prior works \citep{tang2024inference, li2025reneg}, we adopt HPSv2 \citep{wu2023hps} as the reward model since it is a general reward model. For hyperparameters, we set the learning rate $\eta$ to 0.01, 0.02, and 0.06 for SD1.5, SDXL, and SD3 respectively, accounting for differences in model scale. The weighting schedule $w_t$ is computed according to Equation~\ref{eq:scale_weight}. We fix the total number of optimization steps to $S=10$, and set $N=10$ for BoN to ensure a fair comparison. Additional implementation details are provided in Appendix~\ref{sec:app-exp_detail}. We further analyze the trade-off between performance and computational cost in Section~\ref{sec:tradeoff}.

\paragraph{Metrics}
Following prior work \citep{tang2024inference, na2025diffusion}, we primarily use HPSv2 \citep{wu2023hps} as the main metric for evaluating alignment quality. To provide a more comprehensive evaluation, we further report results using additional widely-adopted metrics, including PickScore \citep{kirstain2023pickscore}, CLIPScore \citep{hessel2021clipscore}, VQAScore \citep{lin2024evaluating}, and ImageReward \citep{xu2023imagereward}.

\subsection{Main Results}
\label{sec:main_results}
\begin{table}[htbp]
    \centering
    \small
    \caption{Results on HPDv2 test set. HPSv2 is used as reward model.}
    \begin{tabular}{c|c|ccccc}
    \toprule
       Base Model  & Method & HPSv2 & PickScore & CLIPScore & VQAScore & ImageReward \\
    \midrule
    \multirow{3}{*}{SD3} & Base & 29.71 & 22.33 & 29.46 & 83.94 & 1.105 \\
    ~ & BoN & 32.07 & 22.75 & 29.72 & 84.96 & 1.269 \\
    ~ & DTO(Ours) & \textbf{43.78} & \textbf{24.50} & \textbf{30.14} & \textbf{85.36} & \textbf{1.289}  \\
    \midrule
    \multirow{4}{*}{SDXL} & Base & 28.19	& 22.58 & 30.78 & 82.60 & 0.874\\
    ~ & BoN & 30.90 & 23.10 & 31.06 & 84.33 & 1.155 \\
    ~ & DNO & 38.23	& 24.02	& \textbf{31.09} & 83.77 & 1.145 \\
    ~ & DTO(Ours) & \textbf{43.01} & \textbf{24.87} & 31.04 &  \textbf{84.63} & \textbf{1.220} \\
    \midrule
    \multirow{6}{*}{SD1.5} & Base & 23.48 & 20.72 & 28.98 & 70.36 & 0.015 \\
    ~ & BoN & 27.48 & 21.35 & 30.21 & 76.92 & 0.565\\
    ~ & DNO & 33.70	& 22.29	& 30.21	& 76.66 & 0.591  \\
    ~ & ReNeg & 29.07 & 21.11 & 26.64 & 66.68 & 0.173  \\
    ~ & DATE & 29.19 & 21.38 & 28.47 & 72.54 & 0.339 \\
    ~ & DTO(Ours) & \textbf{40.75} & \textbf{23.41} & \textbf{30.51} & \textbf{79.20} & \textbf{0.702}  \\
    \bottomrule
    \end{tabular}
    \label{tab:exp_hps_hpd_main}
\end{table}

\begin{table}[htbp]
    \centering
    \small
    \caption{Results on GenAIBench. HPSv2 is used as reward model.}
    \begin{tabular}{c|c|ccccc}
    \toprule
       Base Model  & Method & HPSv2 & PickScore & CLIPScore & VQAScore & ImageReward\\
    \midrule
    \multirow{3}{*}{SD3} & Base & 29.12 & 22.50 & 28.96 & 75.61 & 1.065 \\
    ~ & BoN & 31.19 & 22.89 & 29.19 & \textbf{77.03} & 1.260 \\
    ~ & DTO(Ours) & \textbf{41.95} & \textbf{24.58} & \textbf{29.46} & 76.69 & \textbf{1.267} \\
    \midrule
    \multirow{4}{*}{SDXL} & Base & 27.54	& 22.44	& 28.79	& 71.60 & 0.730 \\
    ~ & BoN & 30.24 & 22.99 & \textbf{29.31} & \textbf{73.98} & 1.066 \\
    ~ & DNO & 32.91	& 23.28 & 29.17	& 72.52 & 0.942 \\
    ~ & DTO(Ours) & \textbf{40.57} & \textbf{24.58} & 29.26 & 73.64 & \textbf{1.071} \\
    \midrule
    \multirow{6}{*}{SD1.5} & Base & 24.74 & 21.21 & 27.67 & 61.01 & 0.047 \\
    ~ & BoN & 28.17 & 21.82 & 28.53 & 66.85 & 0.610 \\
    ~ & DNO & 30.20	& 22.16 & 28.25 & 62.55 & 0.414\\
    ~ & ReNeg & 28.09 & 21.44 & 26.38 & 56.55 & 0.112 \\
    ~ & DATE & 27.50 & 21.65 & 27.77 & 61.06 & 0.221 \\
    ~ & DTO(Ours) & \textbf{39.20} & \textbf{23.49} & \textbf{28.95} & \textbf{67.61} & \textbf{0.617} \\
    \bottomrule
    \end{tabular}
    \label{tab:exp_hps_genai_main}
\end{table}

We present the main results on two benchmarks in Tables~\ref{tab:exp_hps_hpd_main} and \ref{tab:exp_hps_genai_main}. Overall, our method DTO outperforms all baselines across three backbone models on both benchmarks and most improvements are statistically significant, with details in Appendix \ref{sec:app-sig}.

\paragraph{Alignment Performance}
We first analyze alignment performance in terms of the primary metric HPSv2. As shown in both tables, DTO achieves substantial improvements over all compared methods across SD1.5, SDXL, and SD3, demonstrating its effectiveness for inference-time alignment of text-to-image diffusion models. In contrast, some existing methods, such as ReNeg and DATE, exhibit limited gains over the base model, indicating suboptimal performance under test-time settings.

\paragraph{Robustness Analysis}
Importantly, the improvements in HPSv2 do not come at the cost of overfitting to the reward model. Across auxiliary evaluation metrics, DTO maintains improvements or competitive performance. In particular, ReNeg even leads to notable degradation compared to the base model on CLIPScore and VQAScore, suggesting potential reward over-optimization or instability.

In contrast, DTO improves or matches strong baselines across most metrics, including PickScore, CLIPScore, VQAScore, and ImageReward. This indicates that our method not only optimizes for the reward model effectively, but also preserves general visual quality and semantic alignment, alleviating concerns of reward hacking and providing genuine improvement. Results that instead use CLIPScore or PickScore as the optimization reward are reported in Appendices~\ref{sec:app-clip} and \ref{sec:app-pick}.

\begin{figure}
    \centering
    \begin{subfigure}[b]{0.45\textwidth}
        \centering
        \includegraphics[width=1.0\linewidth]{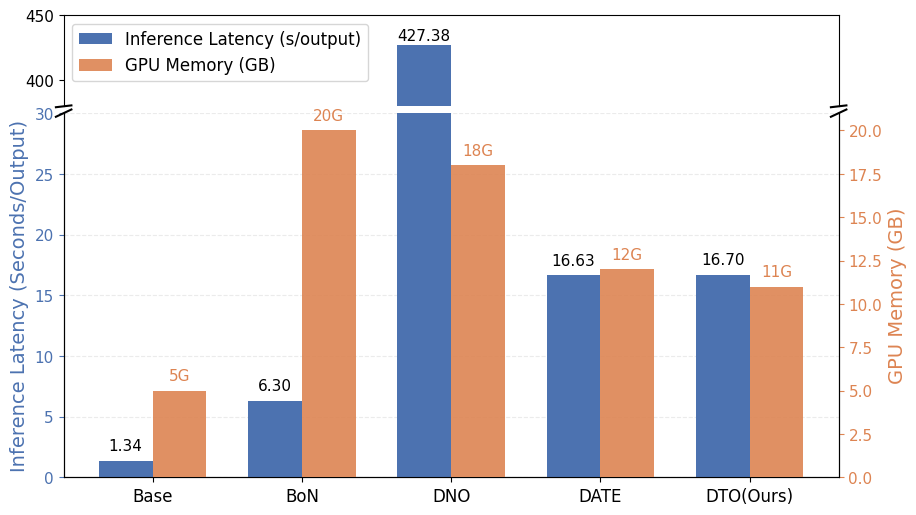}
        \caption{Computation cost comparison of different methods.The computation cost of BoN is calculated using batchified inference ($N$ candidates are generated simultanously), resulting in less inference time yet higher GPU memory cost. SD1.5 is used as the base model.}
        \label{fig:latency}
    \end{subfigure}
    \hfill
    \begin{subfigure}[b]{0.45\textwidth}
        \centering
        \includegraphics[width=1.0\linewidth]{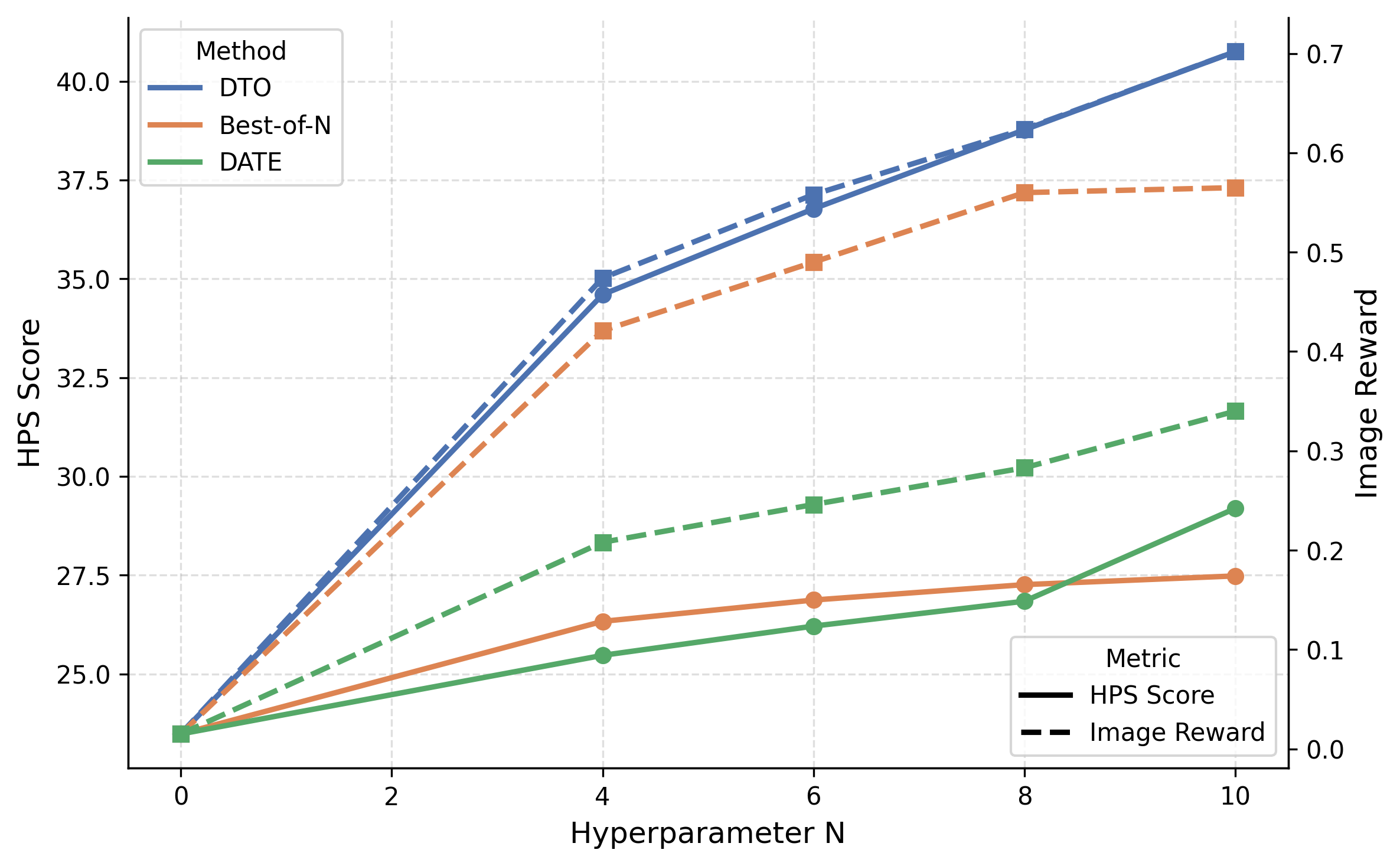}
        \caption{Comparison of scaling behaviour of DTO, BoN and DATE. $N=0$ refers to the base model.}
        \label{fig:scale}
    \end{subfigure}
    \caption{Analysis on efficiency and scaling beahviour of DTO.}
\end{figure}

\subsection{Trade-off Between Performance and Cost}
\label{sec:tradeoff}

We first report the computational cost on an Nvidia H100 GPU in Figure~\ref{fig:latency}, with detailed numbers and a wall-clock-matched comparison against BoN ($N=20$) provided in Appendix \ref{sec:app-latency}. DTO achieves a substantially better efficiency–performance trade-off compared to prior methods. In particular, compared with the strongest optimization-based baseline DNO, DTO achieves approximately a 10$\times$ reduction in inference latency while maintaining higher alignment performance. Compared with more efficient methods such as DATE, DTO exhibits comparable runtime while requiring lower GPU memory, and simultaneously delivers substantially stronger alignment quality. These results demonstrate that DTO provides a more favourable efficiency-effectiveness tradeoff among all inference-time alignment methods.

\paragraph{Scalability analysis}
We further investigate the scaling behavior of DTO. Specifically, we vary the number of optimization steps $S$ for DTO and DATE \footnote{DNO is not discussed due to its extremely long inference time.}, and the number of candidate samples $N$ for BoN (both represented on the x-axis in Figure~\ref{fig:scale}), and report the corresponding HPSv2 and ImageReward trends. As illustrated in the figure, DTO consistently outperforms both BoN and DATE across different hyperparameter settings. Notably, both BoN and DATE exhibit quick saturation in performance, suggesting that their improvement capacity is limited by their respective optimization paradigms. In contrast, DTO demonstrates a more favorable scaling trend: performance continues to improve steadily as the number of optimization steps 
$S$ increases. This indicates that DTO scales more effectively with additional optimization budgets, highlighting its superior scalability compared to prior approaches.

\begin{figure}
    \centering
    \begin{subfigure}[b]{0.4\textwidth}
        \centering
        \includegraphics[width=1.0\linewidth]{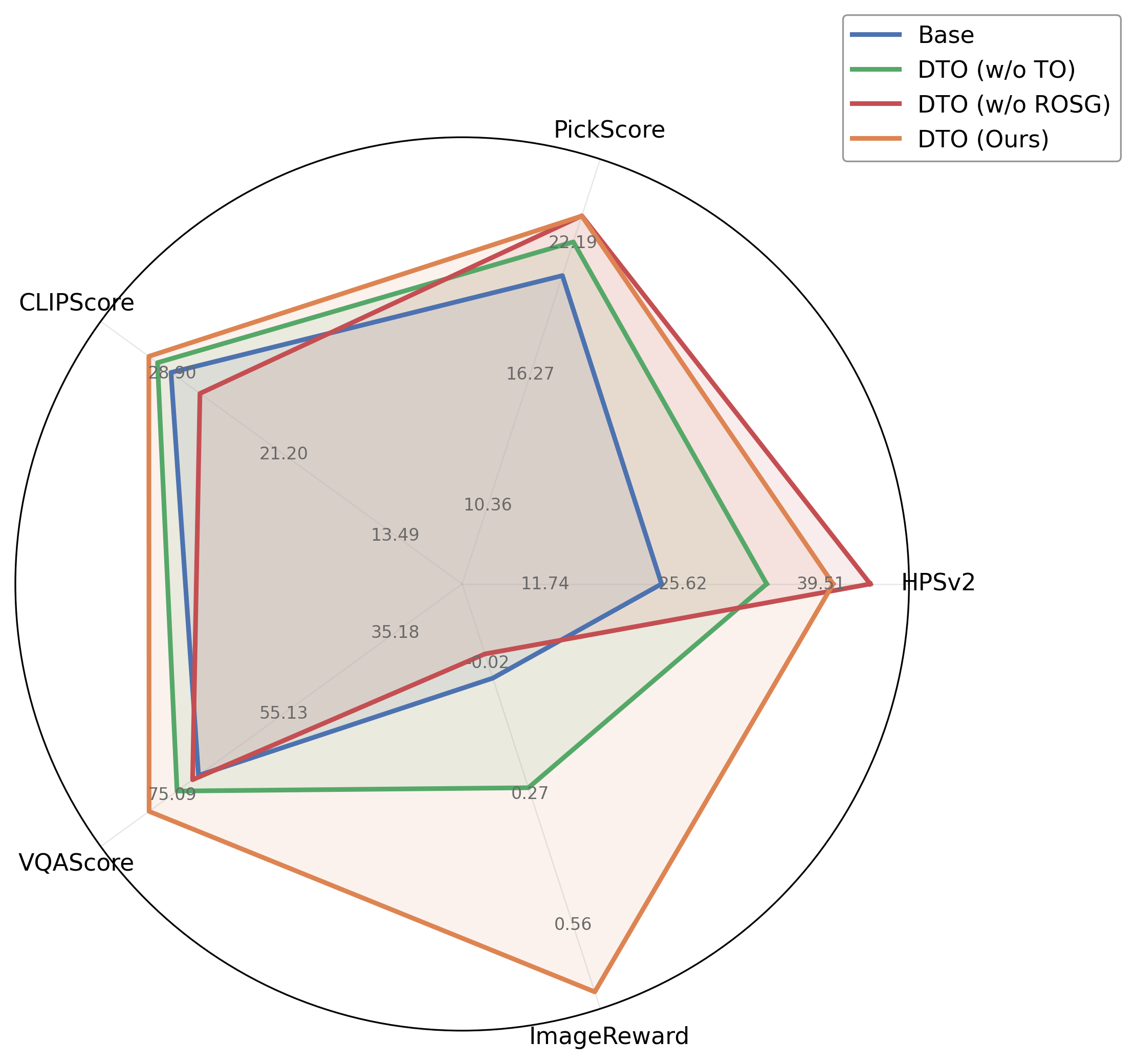}
        \caption{Ablation study.}
        \label{fig:abla}
    \end{subfigure}
    \begin{subfigure}[b]{0.43\textwidth}
        \centering
        \includegraphics[width=1.0\linewidth]{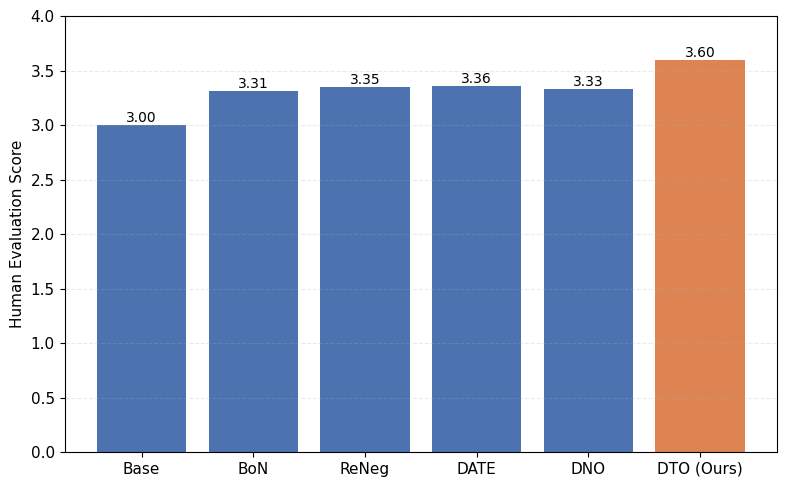}
        \caption{Human evaluation results.}
        \label{fig:human_eval}
    \end{subfigure}
    \caption{Ablation study and human evaluation results.}
\end{figure}

\subsection{Ablation Study}
First, to evaluate the necessity of trajectory-level optimization, we consider an alternative formulation that directly optimizes the final generated image 
$x_0$  via $\max_{x_0} R(x_0)$. We denote this variant as \textit{DTO (w/o TO)}. Next, to study the effect of Reward-Optimized Score Guidance, we set $w_t=1$ for all timesteps. This variant is referred to as \textit{DTO (w/o ROSG)}. Other detailed experiments validating the effectiveness of DTO are shown in Appendix~\ref{sec:app-abla}, including robustness to the learning rate, the score-drift and FID comparison for ROSG, and which trajectory variables are optimized.

The results are reported in Figure~\ref{fig:abla}, with detailed numerical results provided in Appendix~\ref{sec:app-abla}. Due to the restricted optimization space, directly optimizing 
$x_0$  yields suboptimal HPSv2 scores. Removing ROSG, on the other hand, leads to higher reward values but degrades performance on other evaluation metrics, indicating a potential OOD problem. These observations confirm that both full trajectory optimization and Reward-Optimized Score Guidance are essential components of DTO.

\subsection{Human Evaluation and Case Study}
We conduct a human evaluation on the HPDv2 test set using SD1.5 as the base model. For each method, 50 samples generated from the same set of prompts are evaluated. Two annotators are asked to rate each generated image on a 1–5 Likert scale, considering both visual quality and prompt alignment. The results are summarized in Figure~\ref{fig:human_eval}, with numerical scores, annotation criteria, and an additional SD3 evaluation provided in Appendix~\ref{sec:app-human_eval}. The results show that DTO achieves the highest overall preference among all compared methods, which is consistent with the automatic metric results, further validating the effectiveness of DTO.

\begin{figure}
    \centering
    \begin{subfigure}[b]{0.21\textwidth}
        \centering
        \includegraphics[width=1.0\linewidth]{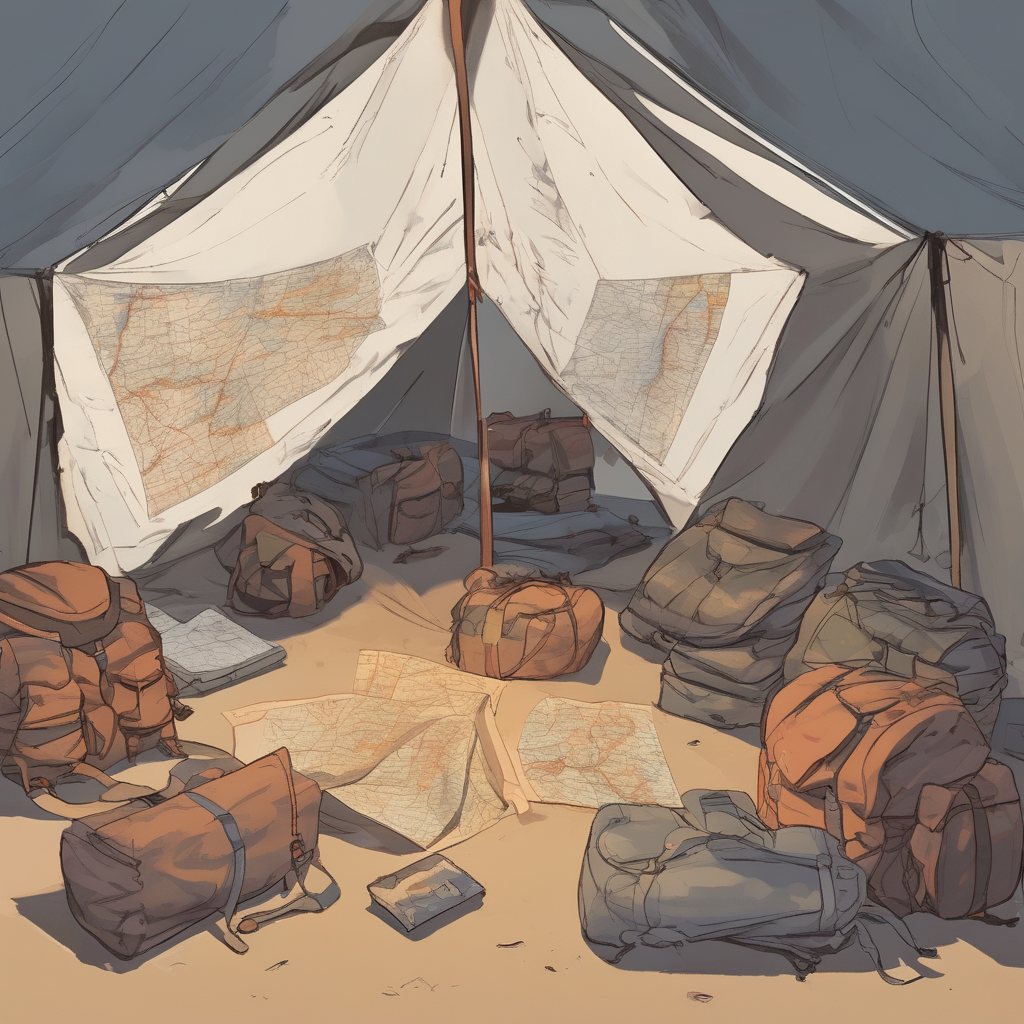}
        \caption{SDXL}
    \end{subfigure}
    \hfill
    \begin{subfigure}[b]{0.21\textwidth}
        \centering
        \includegraphics[width=1.0\linewidth]{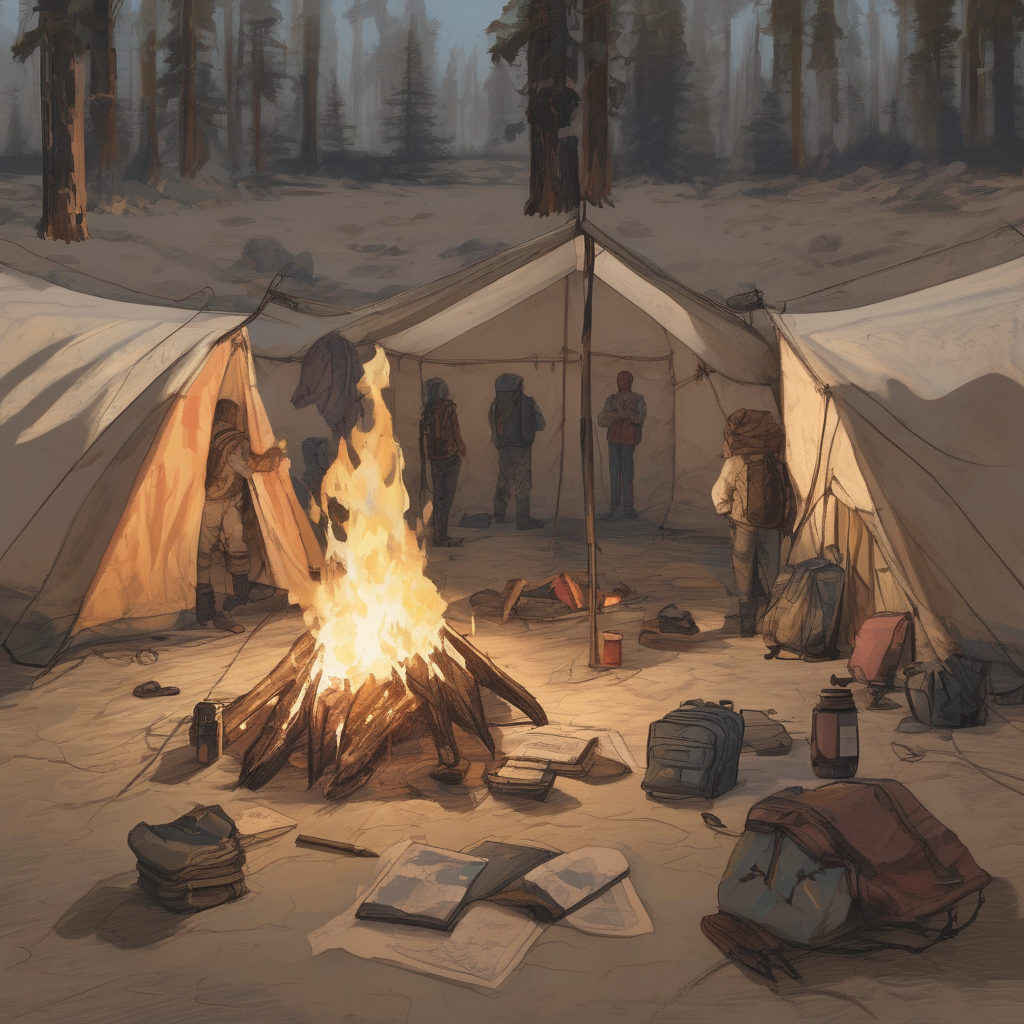}
        \caption{SDXL+BoN}
    \end{subfigure}
    \hfill
    \begin{subfigure}[b]{0.21\textwidth}
        \centering
        \includegraphics[width=1.0\linewidth]{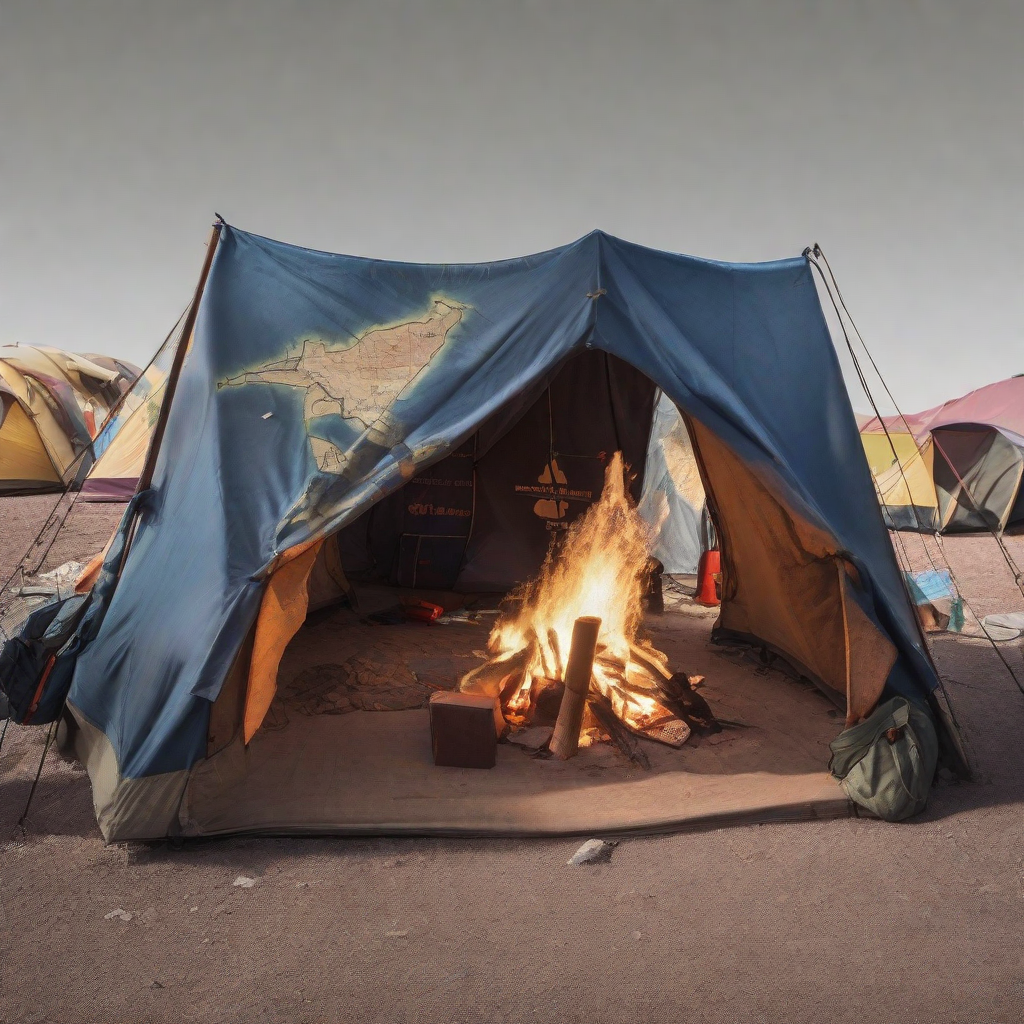}
        \caption{SDXL+DNO \citep{tang2024inference}}
    \end{subfigure}
    \hfill
    \begin{subfigure}[b]{0.21\textwidth}
        \centering
        \includegraphics[width=1.0\linewidth]{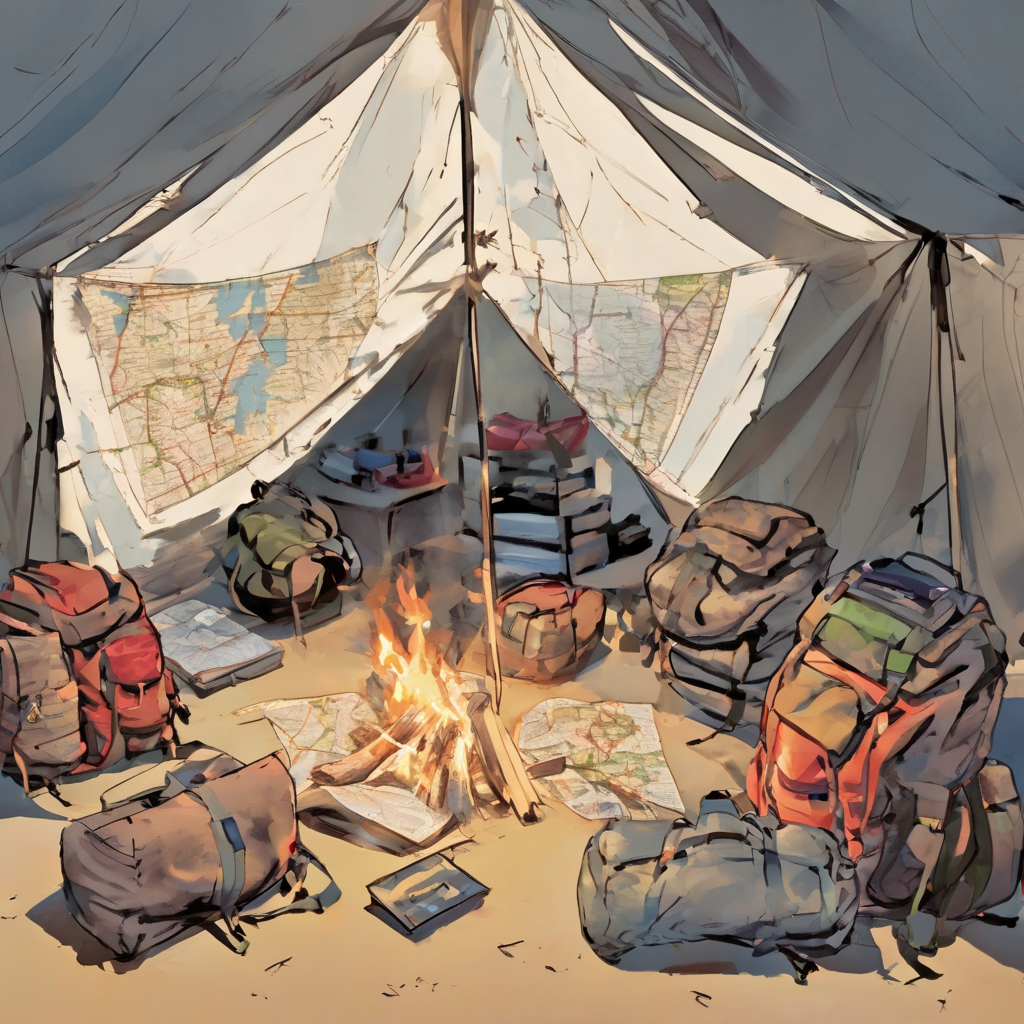}
        \caption{SDXL+DTO(Ours)}
    \end{subfigure}
    \caption{Generation results of different alignment methods using SDXL. The prompt is ``Inside the camp, a fire crackles, casting shadows on the tent walls, with backpacks and maps spread out.''}
     \label{fig:exp_case}
\end{figure}

We further present qualitative comparisons in Figure~\ref{fig:exp_case}. As shown in the example, BoN and DNO, though visually plausible, confuse the relation specified by the prompt (``inside the camp''). The base model even fails to generate a noticeable campfire. In contrast, DTO generates an image that is both more coherent with the text and visually plausible, with richer textures and finer details, demonstrating clear qualitative improvements. Further cases, including how the sample evolves during optimization and a failure case, are provided in Appendix~\ref{sec:app-case}.

\section{Conclusion}

We propose Direct Trajectory Optimization (DTO), a simple yet effective framework for inference-time alignment of text-to-image diffusion models. DTO directly optimizes the diffusion sampling trajectory and incorporates Reward-Optimized Score Guidance to mitigate OOD problem, enabling both efficient optimization and improved alignment performance. Extensive experiments across three models and two benchmarks show that DTO  outperforms existing methods. In future work, we aim to further improve optimization efficiency and extend DTO to non-differentiable reward settings.

\bibliography{iclr2027_conference}

@article{tang2024inference,
  title={Inference-time alignment of diffusion models with direct noise optimization},
  author={Tang, Zhiwei and Peng, Jiangweizhi and Tang, Jiasheng and Hong, Mingyi and Wang, Fan and Chang, Tsung-Hui},
  journal={arXiv preprint arXiv:2405.18881},
  year={2024}
}

@article{kingma2024understanding,
  title={Understanding diffusion objectives as the elbo with simple data augmentation},
  author={Kingma, Diederik and Gao, Ruiqi},
  journal={Advances in Neural Information Processing Systems},
  volume={36},
  year={2024}
}

@misc{krippendorff2011computing,
  title={Computing Krippendorff’s alpha-reliability},
  author={Krippendorff, Klaus},
  year={2011},
  publisher={Citeseer}
}

@article{lu2022dpm,
  title={Dpm-solver: A fast ode solver for diffusion probabilistic model sampling in around 10 steps},
  author={Lu, Cheng and Zhou, Yuhao and Bao, Fan and Chen, Jianfei and Li, Chongxuan and Zhu, Jun},
  journal={Advances in Neural Information Processing Systems},
  volume={35},
  pages={5775--5787},
  year={2022}
}

@misc{zhang2025rethinkingautomaticevaluationimagetext,
      title={Re-Thinking the Automatic Evaluation of Image-Text Alignment in Text-to-Image Models}, 
      author={Huixuan Zhang and Xiaojun Wan},
      year={2025},
      eprint={2506.08480},
      archivePrefix={arXiv},
      primaryClass={cs.CL},
      url={https://arxiv.org/abs/2506.08480}, 
}

@article{karras2022elucidating,
  title={Elucidating the design space of diffusion-based generative models},
  author={Karras, Tero and Aittala, Miika and Aila, Timo and Laine, Samuli},
  journal={Advances in neural information processing systems},
  volume={35},
  pages={26565--26577},
  year={2022}
}

@article{lipman2022flow,
  title={Flow matching for generative modeling},
  author={Lipman, Yaron and Chen, Ricky TQ and Ben-Hamu, Heli and Nickel, Maximilian and Le, Matt},
  journal={arXiv preprint arXiv:2210.02747},
  year={2022}
}

@article{ho2022classifier,
  title={Classifier-free diffusion guidance},
  author={Ho, Jonathan and Salimans, Tim},
  journal={arXiv preprint arXiv:2207.12598},
  year={2022}
}

@inproceedings{esser2024scaling,
  title={Scaling rectified flow transformers for high-resolution image synthesis},
  author={Esser, Patrick and Kulal, Sumith and Blattmann, Andreas and Entezari, Rahim and M{\"u}ller, Jonas and Saini, Harry and Levi, Yam and Lorenz, Dominik and Sauer, Axel and Boesel, Frederic and others},
  booktitle={Forty-first International Conference on Machine Learning},
  year={2024}
}

@article{podell2023sdxl,
  title={Sdxl: Improving latent diffusion models for high-resolution image synthesis},
  author={Podell, Dustin and English, Zion and Lacey, Kyle and Blattmann, Andreas and Dockhorn, Tim and M{\"u}ller, Jonas and Penna, Joe and Rombach, Robin},
  journal={arXiv preprint arXiv:2307.01952},
  year={2023}
}

@article{chen2024pixartsigma,
  title={PixArt-$\Sigma$: Weak-to-Strong Training of Diffusion Transformer for 4K Text-to-Image Generation},
  author={Junsong Chen and Chongjian Ge and Enze Xie and Yue Wu and Lewei Yao and Xiaozhe Ren and Zhongdao Wang and Ping Luo and Huchuan Lu and Zhenguo Li},
  year={2024},
  eprint={2403.04692},
  journal={arXiv},
  primaryClass={cs.CV}
}

@InProceedings{Rombach_2022_CVPR,
    author    = {Rombach, Robin and Blattmann, Andreas and Lorenz, Dominik and Esser, Patrick and Ommer, Bj\"orn},
    title     = {High-Resolution Image Synthesis With Latent Diffusion Models},
    booktitle = {Proceedings of the IEEE/CVF Conference on Computer Vision and Pattern Recognition (CVPR)},
    month     = {June},
    year      = {2022},
    pages     = {10684-10695}
}

@inproceedings{hessel2021clipscore,
  title={CLIPScore: A Reference-free Evaluation Metric for Image Captioning},
  author={Hessel, Jack and Holtzman, Ari and Forbes, Maxwell and Le Bras, Ronan and Choi, Yejin},
  booktitle={Proceedings of the 2021 Conference on Empirical Methods in Natural Language Processing},
  pages={7514--7528},
  year={2021}
}

@inproceedings{NEURIPS2020_DDPM,
 author = {Ho, Jonathan and Jain, Ajay and Abbeel, Pieter},
 booktitle = {Advances in Neural Information Processing Systems},
 editor = {H. Larochelle and M. Ranzato and R. Hadsell and M.F. Balcan and H. Lin},
 pages = {6840--6851},
 publisher = {Curran Associates, Inc.},
 title = {Denoising Diffusion Probabilistic Models},
 url = {https://proceedings.neurips.cc/paper_files/paper/2020/file/4c5bcfec8584af0d967f1ab10179ca4b-Paper.pdf},
 volume = {33},
 year = {2020}
}

@inproceedings{song2021DDIM,
    title={Denoising Diffusion Implicit Models},
    author={Jiaming Song and Chenlin Meng and Stefano Ermon},
    booktitle={International Conference on Learning Representations},
    year={2021},
    url={https://openreview.net/forum?id=St1giarCHLP}
}

@article{lu2022dpm++,
  title={Dpm-solver++: Fast solver for guided sampling of diffusion probabilistic models},
  author={Lu, Cheng and Zhou, Yuhao and Bao, Fan and Chen, Jianfei and Li, Chongxuan and Zhu, Jun},
  journal={arXiv preprint arXiv:2211.01095},
  year={2022}
}

@inproceedings{nichol2021improved,
  title={Improved denoising diffusion probabilistic models},
  author={Nichol, Alexander Quinn and Dhariwal, Prafulla},
  booktitle={International conference on machine learning},
  pages={8162--8171},
  year={2021},
  organization={PMLR}
}

@article{dhariwal2021diffusion,
  title={Diffusion models beat gans on image synthesis},
  author={Dhariwal, Prafulla and Nichol, Alexander},
  journal={Advances in neural information processing systems},
  volume={34},
  pages={8780--8794},
  year={2021}
}

@inproceedings{song2021scorebased,
    title={Score-Based Generative Modeling through Stochastic Differential Equations},
    author={Yang Song and Jascha Sohl-Dickstein and Diederik P Kingma and Abhishek Kumar and Stefano Ermon and Ben Poole},
    booktitle={International Conference on Learning Representations},
    year={2021},
    url={https://openreview.net/forum?id=PxTIG12RRHS}
}

@article{liu2022flow,
  title={Flow straight and fast: Learning to generate and transfer data with rectified flow},
  author={Liu, Xingchao and Gong, Chengyue and Liu, Qiang},
  journal={arXiv preprint arXiv:2209.03003},
  year={2022}
}

@article{albergo2022building,
  title={Building normalizing flows with stochastic interpolants},
  author={Albergo, Michael S and Vanden-Eijnden, Eric},
  journal={arXiv preprint arXiv:2209.15571},
  year={2022}
}

@article{na2025diffusion,
  title={Diffusion Adaptive Text Embedding for Text-to-Image Diffusion Models},
  author={Na, Byeonghu and Park, Minsang and Sim, Gyuwon and Shin, Donghyeok and Bae, HeeSun and Kang, Mina and Kwon, Se Jung and Kang, Wanmo and Moon, Il-Chul},
  journal={arXiv preprint arXiv:2510.23974},
  year={2025}
}

@inproceedings{li2025reneg,
  title={Reneg: Learning negative embedding with reward guidance},
  author={Li, Xiaomin and Liu, Yixuan and Isobe, Takashi and Jia, Xu and Cui, Qinpeng and Zhou, Dong and Li, Dong and He, You and Lu, Huchuan and Wang, Zhongdao and others},
  booktitle={Proceedings of the Computer Vision and Pattern Recognition Conference},
  pages={23636--23645},
  year={2025}
}

@article{wu2023hps,
  title={Human Preference Score v2: A Solid Benchmark for Evaluating Human Preferences of Text-to-Image Synthesis},
  author={Wu, Xiaoshi and Hao, Yiming and Sun, Keqiang and Chen, Yixiong and Zhu, Feng and Zhao, Rui and Li, Hongsheng},
  journal={arXiv preprint arXiv:2306.09341},
  year={2023}
}

@inproceedings{li2024evaluating,
  title={Evaluating and improving compositional text-to-visual generation},
  author={Li, Baiqi and Lin, Zhiqiu and Pathak, Deepak and Li, Jiayao and Fei, Yixin and Wu, Kewen and Xia, Xide and Zhang, Pengchuan and Neubig, Graham and Ramanan, Deva},
  booktitle={Proceedings of the IEEE/CVF Conference on Computer Vision and Pattern Recognition},
  pages={5290--5301},
  year={2024}
}

@misc{kirstain2023pickscore,
      title={Pick-a-Pic: An Open Dataset of User Preferences for Text-to-Image Generation}, 
      author={Yuval Kirstain and Adam Polyak and Uriel Singer and Shahbuland Matiana and Joe Penna and Omer Levy},
      year={2023},
      eprint={2305.01569},
      archivePrefix={arXiv},
      primaryClass={cs.CV},
      url={https://arxiv.org/abs/2305.01569}, 
}

@article{lin2024evaluating,
        title={Evaluating Text-to-Visual Generation with Image-to-Text Generation},
        author={Lin, Zhiqiu and Pathak, Deepak and Li, Baiqi and Li, Jiayao and Xia, Xide and Neubig, Graham and Zhang, Pengchuan and Ramanan, Deva},
        journal={arXiv preprint arXiv:2404.01291},
        year={2024}
      }

@misc{wu2025qwenimagetechnicalreport,
      title={Qwen-Image Technical Report}, 
      author={Chenfei Wu and Jiahao Li and Jingren Zhou and Junyang Lin and Kaiyuan Gao and Kun Yan and Sheng-ming Yin and Shuai Bai and Xiao Xu and Yilei Chen and Yuxiang Chen and Zecheng Tang and Zekai Zhang and Zhengyi Wang and An Yang and Bowen Yu and Chen Cheng and Dayiheng Liu and Deqing Li and Hang Zhang and Hao Meng and Hu Wei and Jingyuan Ni and Kai Chen and Kuan Cao and Liang Peng and Lin Qu and Minggang Wu and Peng Wang and Shuting Yu and Tingkun Wen and Wensen Feng and Xiaoxiao Xu and Yi Wang and Yichang Zhang and Yongqiang Zhu and Yujia Wu and Yuxuan Cai and Zenan Liu},
      year={2025},
      eprint={2508.02324},
      archivePrefix={arXiv},
      primaryClass={cs.CV},
      url={https://arxiv.org/abs/2508.02324}, 
}

@misc{seedream2025seedream40nextgenerationmultimodal,
      title={Seedream 4.0: Toward Next-generation Multimodal Image Generation}, 
      author={Team Seedream and : and Yunpeng Chen and Yu Gao and Lixue Gong and Meng Guo and Qiushan Guo and Zhiyao Guo and Xiaoxia Hou and Weilin Huang and Yixuan Huang and Xiaowen Jian and Huafeng Kuang and Zhichao Lai and Fanshi Li and Liang Li and Xiaochen Lian and Chao Liao and Liyang Liu and Wei Liu and Yanzuo Lu and Zhengxiong Luo and Tongtong Ou and Guang Shi and Yichun Shi and Shiqi Sun and Yu Tian and Zhi Tian and Peng Wang and Rui Wang and Xun Wang and Ye Wang and Guofeng Wu and Jie Wu and Wenxu Wu and Yonghui Wu and Xin Xia and Xuefeng Xiao and Shuang Xu and Xin Yan and Ceyuan Yang and Jianchao Yang and Zhonghua Zhai and Chenlin Zhang and Heng Zhang and Qi Zhang and Xinyu Zhang and Yuwei Zhang and Shijia Zhao and Wenliang Zhao and Wenjia Zhu},
      year={2025},
      eprint={2509.20427},
      archivePrefix={arXiv},
      primaryClass={cs.CV},
      url={https://arxiv.org/abs/2509.20427}, 
}

@inproceedings{zhang-wan-2025-r,
    title = "{R}-Bind: Unified Enhancement of Attribute and Relation Binding in Text-to-Image Diffusion Models",
    author = "Zhang, Huixuan  and
      Wan, Xiaojun",
    editor = "Christodoulopoulos, Christos  and
      Chakraborty, Tanmoy  and
      Rose, Carolyn  and
      Peng, Violet",
    booktitle = "Proceedings of the 2025 Conference on Empirical Methods in Natural Language Processing",
    month = nov,
    year = "2025",
    address = "Suzhou, China",
    publisher = "Association for Computational Linguistics",
    url = "https://aclanthology.org/2025.emnlp-main.349/",
    doi = "10.18653/v1/2025.emnlp-main.349",
    pages = "6856--6870",
    ISBN = "979-8-89176-332-6",
}

@misc{gu2024diffusionrpoaligningdiffusionmodels,
      title={Diffusion-RPO: Aligning Diffusion Models through Relative Preference Optimization}, 
      author={Yi Gu and Zhendong Wang and Yueqin Yin and Yujia Xie and Mingyuan Zhou},
      year={2024},
      eprint={2406.06382},
      archivePrefix={arXiv},
      primaryClass={cs.CV},
      url={https://arxiv.org/abs/2406.06382}, 
}

@misc{ddpo,
      title={Training Diffusion Models with Reinforcement Learning}, 
      author={Kevin Black and Michael Janner and Yilun Du and Ilya Kostrikov and Sergey Levine},
      year={2024},
      eprint={2305.13301},
      archivePrefix={arXiv},
      primaryClass={cs.LG},
      url={https://arxiv.org/abs/2305.13301}, 
}

@misc{wallace2023diffusiondpo,
      title={Diffusion Model Alignment Using Direct Preference Optimization}, 
      author={Bram Wallace and Meihua Dang and Rafael Rafailov and Linqi Zhou and Aaron Lou and Senthil Purushwalkam and Stefano Ermon and Caiming Xiong and Shafiq Joty and Nikhil Naik},
      year={2023},
      eprint={2311.12908},
      archivePrefix={arXiv},
      primaryClass={cs.CV},
      url={https://arxiv.org/abs/2311.12908}, 
}

@misc{flux2024,
    author={Black Forest Labs},
    title={FLUX},
    year={2024},
    howpublished={\url{https://github.com/black-forest-labs/flux}},
}

@article{gao2025seedream,
  title={Seedream 3.0 technical report},
  author={Gao, Yu and Gong, Lixue and Guo, Qiushan and Hou, Xiaoxia and Lai, Zhichao and Li, Fanshi and Li, Liang and Lian, Xiaochen and Liao, Chao and Liu, Liyang and others},
  journal={arXiv preprint arXiv:2504.11346},
  year={2025}
}

@inproceedings{xu2023imagereward,
  title={ImageReward: learning and evaluating human preferences for text-to-image generation},
  author={Xu, Jiazheng and Liu, Xiao and Wu, Yuchen and Tong, Yuxuan and Li, Qinkai and Ding, Ming and Tang, Jie and Dong, Yuxiao},
  booktitle={Proceedings of the 37th International Conference on Neural Information Processing Systems},
  pages={15903--15935},
  year={2023}
}

@article{unifiedreward,
  title={Unified reward model for multimodal understanding and generation},
  author={Wang, Yibin and Zang, Yuhang and Li, Hao and Jin, Cheng and Wang, Jiaqi},
  journal={arXiv preprint arXiv:2503.05236},
  year={2025}
}

@article{unifiedreward-think,
  title={Unified multimodal chain-of-thought reward model through reinforcement fine-tuning},
  author={Wang, Yibin and Li, Zhimin and Zang, Yuhang and Wang, Chunyu and Lu, Qinglin and Jin, Cheng and Wang, Jiaqi},
  journal={arXiv preprint arXiv:2505.03318},
  year={2025}
}

@article{unifiedreward-flex,
  title={Unified Personalized Reward Model for Vision Generation},
  author={Wang, Yibin and Zang, Yuhang and Han, Feng and Bu, Jiazi and Zhou, Yujie and Jin, Cheng and Wang, Jiaqi},
  journal={arXiv preprint arXiv:2602.02380},
  year={2026}
}

@misc{zhang2025minosmultimodalevaluationmodel,
      title={Minos: A Multimodal Evaluation Model for Bidirectional Generation Between Image and Text}, 
      author={Junzhe Zhang and Huixuan Zhang and Xinyu Hu and Li Lin and Mingqi Gao and Shi Qiu and Xiaojun Wan},
      year={2025},
      eprint={2506.02494},
      archivePrefix={arXiv},
      primaryClass={cs.CL},
      url={https://arxiv.org/abs/2506.02494}, 
}

@article{christiano2017deep,
  title={Deep reinforcement learning from human preferences},
  author={Christiano, Paul F and Leike, Jan and Brown, Tom and Martic, Miljan and Legg, Shane and Amodei, Dario},
  journal={Advances in neural information processing systems},
  volume={30},
  year={2017}
}

@article{ouyang2022training,
  title={Training language models to follow instructions with human feedback},
  author={Ouyang, Long and Wu, Jeffrey and Jiang, Xu and Almeida, Diogo and Wainwright, Carroll and Mishkin, Pamela and Zhang, Chong and Agarwal, Sandhini and Slama, Katarina and Ray, Alex and others},
  journal={Advances in neural information processing systems},
  volume={35},
  pages={27730--27744},
  year={2022}
}

@article{rafailov2023direct,
  title={Direct preference optimization: Your language model is secretly a reward model},
  author={Rafailov, Rafael and Sharma, Archit and Mitchell, Eric and Manning, Christopher D and Ermon, Stefano and Finn, Chelsea},
  journal={Advances in neural information processing systems},
  volume={36},
  pages={53728--53741},
  year={2023}
}

@inproceedings{zhang2025much,
  title={How Much To Guide: Revisiting Adaptive Guidance in Classifier-Free Guidance Text-to-Vision Diffusion Models},
  author={Zhang, Huixuan and Wan, Xiaojun},
  booktitle={Proceedings of the 7th ACM International Conference on Multimedia in Asia},
  pages={1--8},
  year={2025}
}
\bibliographystyle{iclr2027_conference}

\appendix
\section{Theoretical Details}

\subsection{Details of Gradient Flow}
\label{sec:app-gf}

\paragraph{Gradient Flow of DTO}
We present details of the gradient $\dfrac{\partial R}{\partial \tau}$ as follows:

Firstly, since $R$ is differentiable, we can calculate $\dfrac{\partial R(x_{0}, c)}{\partial x_{0}}$ as the initial gradient. Then, since we have $x_{t-1}=F(x_{t}, \epsilon_{t}, t)$, we can iteratively calculate:

\begin{align}
    \dfrac{\partial R}{\partial \epsilon_{t}} = \dfrac{\partial R}{\partial x_{t-1}}\dfrac{\partial x_{t-1}}{\partial \epsilon_{t}} = \dfrac{\partial R}{\partial x_{t-1}}\dfrac{\partial F(x_{t}, \epsilon_{t}, t)}{\partial \epsilon_{t}} \\
    \dfrac{\partial R}{\partial x_{t}} = \dfrac{\partial R}{\partial x_{t-1}}\dfrac{\partial x_{t-1}}{\partial x_{t}} = \dfrac{\partial R}{\partial x_{t-1}}\dfrac{\partial F(x_{t}, \epsilon_{t}, t)}{\partial x_{t}}
    \label{eq:gf_x_dto}
\end{align}

Therefore, the gradient $\dfrac{\partial R}{\partial \tau}$ is:

\begin{equation}
    \dfrac{\partial R}{\partial \tau} = \left(\dfrac{\partial R}{\partial x_{T}}, \dfrac{\partial R}{\partial \epsilon_{T}}, ..., \dfrac{\partial R}{\partial \epsilon_{1}}\right)
\end{equation}

As an example, suppose the noise scheduler follows such a formulation:

\begin{equation}
    x_{t-1} =\alpha_{t}x_{t}+\beta_{t}\epsilon_{t}
\end{equation}

which is often the case, then we have:

\begin{equation}
    x_{0}=f(x_{T}, \epsilon_{T},...,\epsilon_{1})=(\prod_{t=1}^{T}\alpha_{t})x_{T}+\sum_{i=1}^{T}(\prod_{j=1}^{i-1}\alpha_{j})\beta_{i}\epsilon_{i}
\end{equation}

Thus we have:

\begin{equation}
    \dfrac{\partial R}{\partial \epsilon_{t}} = \dfrac{\partial R}{\partial x_{0}}\dfrac{\partial x_{0}}{\partial \epsilon_{t}} = \dfrac{\partial R}{\partial x_{0}}(\prod_{j=1}^{t-1}\alpha_{j})\beta_{t}
\end{equation}

and

\begin{equation}
    \dfrac{\partial R}{\partial x_{T}} = \dfrac{\partial R}{\partial x_{0}}\dfrac{\partial x_{0}}{\partial x_{T}} = \dfrac{\partial R}{\partial x_{0}}(\prod_{j=1}^{T}\alpha_{j})
\end{equation}

\paragraph{Comparing Gradient Flow with Other Methods}

In contrast, previous methods all rely on model parameter $\theta$, resulting in a gradient flow similar to follows:

\begin{equation}
    \dfrac{\partial R}{\partial x_{t}} = \dfrac{\partial R}{\partial x_{t-1}}\dfrac{\partial x_{t-1}}{\partial x_{t}} = \dfrac{\partial R}{\partial x_{t-1}}\left(\dfrac{\partial F(x_{t}, \epsilon_{t}, t)}{\partial x_{t}} + \dfrac{\partial F(x_{t}, \epsilon_{t}, t)}{\partial \epsilon_{t}}\dfrac{\partial \tilde{\epsilon}_{\theta}(x_{t}, t, c, n)}{\partial x_{t}} \right)
    \label{eq:gf_x_other}
\end{equation}

Comparing Equation \ref{eq:gf_x_dto} and \ref{eq:gf_x_other}, a major difference is \ref{eq:gf_x_other} requires calculating $\frac{\partial \tilde{\epsilon}_{\theta}(x_{t}, t, c, n)}{\partial x_{t}}$, which is a back-propagation through the whole denoising network, inducing high inference cost and potential difficulties in optimization. 

For DNO \citep{tang2024inference}, this back-propagation is conducted through the whole denoising process, introducing a repeated calculation of $\frac{\partial \tilde{\epsilon}_{\theta}(x_{t}, t, c, n)}{\partial x_{t}}$, resulting in extremely high computation cost. DATE \citep{na2025diffusion} and ReNeg \citep{li2025reneg} relax this cost by applying a one-step approximation at a certain timestep $t$:

\begin{equation}
    x_{0} = M(x_{T}, c, n, \{z_{t}\}) \approx \tilde{x}_{0} = g(x_{t},\epsilon_{t},t)
\end{equation}

where $g$ is an approximation function determined by noise scheduler $F$, and the gradient is only computed on the last denoising step, such as calculating the gradient on $n$:

\begin{equation}
    \dfrac{\partial R(x_{0},c)}{\partial n} \approx \dfrac{\partial R(\tilde{x}_{0},c)}{\partial n} = \dfrac{\partial R(\tilde{x_0}, c)}{\partial \tilde{x}_{0}}\dfrac{\partial g(x_{t}, \epsilon_{t},t)}{\partial n} = \dfrac{\partial R(\tilde{x_0}, c)}{\partial \tilde{x}_{0}}\dfrac{\partial g(x_{t}, \epsilon_{t},t)}{\partial \epsilon_{t}} \dfrac{\partial \tilde{\epsilon}_{\theta}(x_t, t, c, n)}{\partial n}
\end{equation}

Though this strategy relaxes the requirement to back-propagate through the whole denoising process, it still requires a one-step back-propagation through the denoising network. Also, this approximation is often flawed, leading to more difficulties in optimization and a higher requirement for hyper-parameter selection (as the appropriate timestep $t$ to perform such approximation is non-trivial to select \citep{li2025reneg}).

\paragraph{Exactness of the Trajectory Gradient}
The gradient used by DTO is the exact gradient of the trajectory-level objective $R(f(\tau))$, rather than a stop-gradient surrogate of the denoiser-constrained objective used by prior methods. Under Equations~\ref{eq:dto}--\ref{eq:dto_grad}, the intermediate scores $\epsilon_t$ are free trajectory variables, and $\partial R / \partial \epsilon_t$ is obtained by differentiating through the noise scheduler only. The denoiser is used to initialize $\tau$ and, after each ascent step, to provide the pretrained prediction $\epsilon'_t$ for ROSG; it is not treated as a hard constraint during the trajectory update. This gradient therefore differs from the full analytical gradient through $\tilde{\epsilon}_\theta$ by design: DTO solves trajectory-level optimization rather than approximating the denoiser-constrained problem. Unconstrained updates of $\epsilon_t$ can still drift out of distribution; ROSG interpolates the optimized scores with pretrained predictions after each step and keeps the trajectory in a more stable regime, as quantified in Section~\ref{sec:app-abla}. 

\subsection{SDE-Based DTO}
\label{sec:app-sde}

For SDE-based noise schedulers, the denoising iteration follows:

\begin{equation}
    x_{t-1}=F(x_{t},\epsilon_{t}, t, z_{t})
\end{equation}

We can similarly define a trajectory as:

\begin{equation}
    \tau = (x_{T}, \epsilon_{T}, z_{T}, ...,\epsilon_{1}, z_{1})
\end{equation}

where $x_{T}, \{\epsilon_{t}\}_{t=1}^{T}, \{z_{t}\}_{t=1}^{T}$ are also conceptually independent. The optimization target can be given as:

\begin{equation}
\max_{x_T, \epsilon_T, z_{T}, ..., \epsilon_1,z_1} R = R(f(\tau)) = R(f(x_T, \epsilon_T, z_{T}, ..., \epsilon_1, z_1), c)
\label{eq:dto_sde}
\end{equation}

The gradient of $z_{t}$ can be computed similarly to Equation \ref{eq:dto_grad}, following:
\begin{equation}
    \frac{\partial R}{\partial z_t} = \frac{\partial R}{\partial x_{t-1}} \frac{\partial F(x_t, \epsilon_t, t, z_{t})}{\partial z_t}
\end{equation}

Formally, we note:

\begin{equation}
    \nabla_{\tau}R(f(\tau))=\dfrac{\partial R}{\partial \tau} = \left(\dfrac{\partial R}{\partial x_{T}}, \dfrac{\partial R}{\partial \epsilon_{T}}, \dfrac{\partial R}{\partial z_{T}}, ..., \dfrac{\partial R}{\partial \epsilon_{1}},  \dfrac{\partial R}{\partial z_{1}}\right)
\end{equation}

Other design actually remain the same as ODE-based DTO detailed in the main content.

\subsection{Proof of Theorem 1}
\label{sec:app-proof}

Let $\tau^{'} = \tau +\eta \nabla_{\tau}R(f(\tau))$, applying a one-order Taylor expansion, we have:

\begin{equation*}
    R(f(\tau')) = R(f(\tau))+\nabla_{\tau}R(f(\tau))(\tau'-\tau)+O(||\tau'-\tau||^{2})
\end{equation*}

note that $\tau'-\tau = \eta \nabla_{\tau}R(f(\tau))$, which indicates:

\[
    R(f(\tau'))=R(f(\tau))+\eta ||\nabla_{\tau}R(f(\tau))||^{2}+O(\eta^{2})
\]

Directly applying the definition of Lipschitz Continuity and the absolute inequality, we have:

\[
    R(f(\hat{\tau}))\geq R(f(\tau'))-L||\hat{\tau}-\tau'||
\]

given the definition of L-Lipschitz Continuity that $|R(f(\tau_{1}))-R(f(\tau_{2}))|\leq L||\tau_{1}-\tau_{2}||$.

Following this, we also have:

\[
    ||\hat{\tau}-\tau'||\leq \sum_{t=1}^{T}||\hat{\epsilon_{t}}-\epsilon_{t}|| = \sum_{t=1}^{T}||w_{t}\epsilon_{t}+(1-w_{t})\epsilon'_{t}-\epsilon_{t}|| = \sum_{t=1}^{T}||(1-w_{t})\epsilon_{t}'-(1-w_{t})\epsilon_{t}||
\]

Combining these, we have:

\begin{equation*}
\begin{split}
    R(f(\hat{\tau})) &\geq R(f(\tau'))-L\sum_{t=1}^{T}(1-w_{t})||\epsilon_{t}'-\epsilon_{t}|| \\
     &= R(f(\tau))+\eta ||\nabla_{\tau}R(f(\tau))||^{2} -L\sum_{t=1}^{T}(1-w_{t})||\epsilon_{t}'-\epsilon_{t}|| +O(\eta^{2})
\end{split}
\end{equation*}

\paragraph{Scope of the Bound}
Theorem~\ref{theo:combine} assumes that the composed reward $R(f(\tau))$ is $L$-Lipschitz in the trajectory variables. Under a first-order expansion of a gradient-ascent step, it lower-bounds the reward after ROSG by a positive term proportional to $\|\nabla_{\tau} R\|^2$ and a negative penalty proportional to $\sum_t (1-w_t)\|\epsilon'_t - \epsilon_t\|$. Just like we have mentioned earlier, the guarantee of reward improvement is meaningful only when this deviation term does not dominate the gradient term, up to higher-order terms in the step size $\eta$.

The bound becomes vacuous in two cases. If the optimized scores drift far from the pretrained predictions, the negative term can outweigh $\eta\|\nabla_{\tau} R\|^2$, and the theorem no longer ensures improvement. Likewise, when the reward gradient is already small, the first-order gain is weak and further updates bring limited benefit. Both cases match the discussion following Theorem~\ref{theo:combine}: large deviations indicate an out-of-distribution regime where unconstrained optimization can be counterproductive, which is the failure mode ROSG is designed to mitigate. In practice, ROSG and a moderate step size keep updates within a locally well-behaved region. 

We would like to note that the theorem does not claim that a small $\|\epsilon_t - \epsilon'_t\|$ implies the trajectory is fully in-distribution, nor does it by itself prove that ROSG eliminates OOD. It characterizes the operating regime in which reward improvement remains reliable, while preventing severe OOD is left to the design of ROSG and is validated empirically in Section~\ref{sec:app-abla}.

\section{Experiment Details}

\subsection{Details of Experiment Setup}
\label{sec:app-exp_detail}

We largely follow the default configurations of prior methods for a fair comparison. For ReNeg \citep{li2025reneg}, there are two versions of implementations in the original paper: a pretrained one and an inference-time alignment one. We adopt pretrained negative embeddings since the official implementation of inference-time optimization is not publicly available. As a result, we do not report its computational cost due to the lack of a fair and consistent comparison setting.

For DATE \citep{na2025diffusion}, we perform optimization every 5 denoising steps and set the total number of optimization steps to 10. This configuration is chosen because the default setup in the original work yields limited improvement under our evaluation protocol.

For other inference hyperparameters, the classifier-free guidance (CFG) scales are set to $w=5, 5, 7$, and the number of denoising steps are $T=50, 50, 28$ for Stable-Diffusion-1.5, Stable-Diffusion-XL, and Stable-Diffusion-3, respectively, following commonly adopted settings. The random seed is fixed to 42. For each prompt, we generate a single image, and ensure that all methods using the same base model share the identical initial noise sample, enabling a fair and controlled comparison. Precision is fixed as bfloat16 for all experiments.

\subsection{Significance Verification}
\label{sec:app-sig}
We discuss the significance of the results. We conduct a paired t-test on the HPDv2 evaluation results to verify whether the observed metric improvement of DTO is significant following \citep{zhang2025rethinkingautomaticevaluationimagetext}. We present p-values in Table \ref{tab:exp_sig}.

\begin{table}[htbp]
    \centering
    \caption{p-value on HPDv2 test set evaluating whether DTO is significantly better regarding the given metric. A value less than $0.05$ is labeled in \textbf{bold} to indicate significance (DTO is statistically better). HPSv2 is used as reward model.}
    \begin{tabular}{c|c|ccccc}
    \toprule
       Base Model  & Method & HPSv2 & PickScore & CLIPScore & VQAScore & ImageReward \\
    \midrule
    \multirow{2}{*}{SD3} & Base & \textbf{0.000} & \textbf{0.000} & \textbf{0.000} & \textbf{0.000} & \textbf{0.000} \\
    ~ & BoN & \textbf{0.000} & \textbf{0.000} & \textbf{0.001} & 0.191 & 0.174 \\
    \midrule
    \multirow{3}{*}{SDXL} & Base & \textbf{0.000} & \textbf{0.000} & \textbf{0.000} & \textbf{0.000} & \textbf{0.000} \\
    ~ & BoN & \textbf{0.000} & \textbf{0.000} & 0.572 & 0.298 & \textbf{0.006} \\
    ~ & DNO & \textbf{0.000}	& \textbf{0.000}	& 0.672 & 0.108 & \textbf{0.008} \\
    \midrule
    \multirow{5}{*}{SD1.5} & Base & \textbf{0.000} & \textbf{0.000} & \textbf{0.000} & \textbf{0.000} & \textbf{0.000}\\
    ~ & BoN & \textbf{0.000} & \textbf{0.000} & \textbf{0.033} & \textbf{0.005} & \textbf{0.000} \\
    ~ & DNO & \textbf{0.000}	& \textbf{0.000}	& \textbf{0.033}	& \textbf{0.005} & \textbf{0.006}  \\
    ~ & ReNeg & \textbf{0.000} & \textbf{0.000} & \textbf{0.000} & \textbf{0.000} & \textbf{0.000}  \\
    ~ & DATE & \textbf{0.000} & \textbf{0.000} & \textbf{0.000} & \textbf{0.000} & \textbf{0.000} \\
    \bottomrule
    \end{tabular}
    \label{tab:exp_sig}
\end{table}

As can be seen, most results show statistical significance, indicating that the observed improvement does not come from random rare cases. For the rest few non-significant cases, DTO still maintains competitive performance with a p-value less than $0.7$, highlighting DTO bears stable performance.

\subsection{Additional Results Using CLIPScore as Reward Model}
\label{sec:app-clip}
We present results using CLIPScore as reward model in Table \ref{tab:exp_clip_hpd_main}. 

\begin{table}[htbp]
    \centering
    \caption{Results on HPDv2 test set. CLIPScore is used as reward model.}
    \begin{tabular}{c|c|ccccc}
    \toprule
       Base Model  & Method & HPSv2 & PickScore & CLIPScore & VQAScore & ImageReward \\
    \midrule
    \multirow{3}{*}{SD3} & Base & 29.71 & 22.33 & 29.46 & 83.94 & 1.105 \\
    ~ & BoN & \textbf{30.05} & \textbf{22.49} & 32.04 & 87.35 & \textbf{1.216} \\
    ~ & DTO(Ours) & 29.66 & 22.32 & \textbf{45.62} & \textbf{88.60} & 1.128\\
    \midrule
    \multirow{4}{*}{SDXL} & Base & 28.19	& 22.58 & 30.78 & 82.60 & 0.874 \\
    ~ & BoN & 28.31 & \textbf{22.66} & 33.29 & 85.40 & 0.944 \\
    ~ & DNO & \textbf{28.68} & 22.62	& 32.66	& 83.90 & \textbf{0.945} \\
    ~ & DTO(Ours) & 28.40 & 22.54 & \textbf{44.46} & \textbf{87.33} & \textbf{0.945} \\
    \midrule
    \multirow{5}{*}{SD1.5} & Base & 23.48 & 20.72 & 28.98 & 70.36 & 0.015 \\
    ~ & BoN & 24.67 & \textbf{21.04} & 32.71 & 79.47 & \textbf{0.351} \\
    ~ & DNO & 24.20 & 20.89 & 31.91 & 75.45 & 0.159 \\
    ~ & DATE & 23.96 & 20.82 & 31.28 & 74.52 & 0.069 \\
    ~ & DTO(Ours) & \textbf{24.89} & 20.82 & \textbf{44.56} & \textbf{83.14} & 0.241 \\
    \bottomrule
    \end{tabular}
    \label{tab:exp_clip_hpd_main}
\end{table}

CLIPScore is a specialized metric that primarily measures image–text alignment; therefore, optimizing with this reward places greater emphasis on improving semantic consistency between generated images and prompts. As shown in the results, DTO also achieves the highest VQAScore across all three base models, further demonstrating its strong capability in capturing image–text alignment. Moreover, DTO remains competitive on other evaluation metrics, indicating that this improvement is not a reward hacking as it maintains reasonable overall quality.

\subsection{Additional Results Using PickScore as Reward Model}
\label{sec:app-pick}
To further test sensitivity to the choice of reward model, we replace the optimization objective with PickScore on SD3, keeping the remaining settings the same as in Section~\ref{sec:main_results} (including BoN with $N=10$). Results on the HPDv2 test set are reported in Table~\ref{tab:exp_pick_hpd}.

\begin{table}[htbp]
    \centering
    \caption{Results on HPDv2 test set. PickScore is used as the reward model. SD3 is used as the base model.}
    \begin{tabular}{c|ccccc}
    \toprule
      Method & HPSv2 & PickScore & CLIPScore & VQAScore & ImageReward \\
    \midrule
    Base & 29.71 & 22.33 & 29.46 & 83.94 & 1.105 \\
    BoN & 31.10 & 22.99 & 29.96 & 85.30 & \textbf{1.257} \\
    DTO(Ours) & \textbf{35.29} & \textbf{28.53} & \textbf{30.43} & \textbf{85.58} & 1.223 \\
    \bottomrule
    \end{tabular}
    \label{tab:exp_pick_hpd}
\end{table}

DTO substantially improves the optimized PickScore over both the base model and BoN, and also improves the held-out metrics HPSv2, CLIPScore, and VQAScore. ImageReward remains competitive with BoN. These results suggest that DTO remains effective when the reward model is changed.

\subsection{Additional Results on More Backbones}
\label{sec:app-backbones}
We further evaluate DTO on a few-step distilled sampler and a larger DiT-based backbone. Unless otherwise stated, the evaluation protocol follows Section~\ref{sec:main_results}, and HPSv2 is used as the reward model.

\paragraph{Distilled Sampler}
We evaluate DTO on Stable-Diffusion-3.5-Large-Turbo \citep{esser2024scaling} with 4-step sampling, keeping the remaining setup the same as SD3 in the main experiments. Results are reported in Table~\ref{tab:exp_sd35_turbo}.

\begin{table}[htbp]
    \centering
    \caption{Results on HPDv2 test set using Stable-Diffusion-3.5-Large-Turbo with 4-step sampling. HPSv2 is used as the reward model.}
    \begin{tabular}{c|ccccc}
    \toprule
      Method & HPSv2 & PickScore & CLIPScore & VQAScore & ImageReward  \\
    \midrule
    Base & 30.22 & 22.87 & 29.33 & 83.09 & 1.097  \\
    BoN & 32.34 & 23.25 & 29.73 & 84.31 & 1.293 \\
    DTO(Ours) & \textbf{45.86} & \textbf{25.31} & \textbf{30.06} & \textbf{84.74} & \textbf{1.372} \\
    \bottomrule
    \end{tabular}
    \label{tab:exp_sd35_turbo}
\end{table}

Even with only a few denoising steps, DTO still brings gains over both the base model and BoN across all reported metrics. This suggests that DTO remains effective when the sampling trajectory is short. As expected for inference-time optimization, DTO increases latency. When the base model becomes faster, DTO also becomes correspondingly faster, because each optimization iteration still relies on the underlying sampler and ROSG. In regimes where minimal latency is the primary goal, standard or distilled sampling alone is more appropriate. DTO is intended for settings where alignment quality is prioritized over inference efficiency.

\paragraph{FLUX.1-dev}
To examine whether DTO scales to a larger modern backbone, we evaluate it on FLUX.1-dev. All hyperparameters remain the same as in the SD3 experiments. Results are reported in Table~\ref{tab:exp_flux}. 

\begin{table}[htbp]
    \centering
    \caption{Results on HPDv2 test set using FLUX.1-dev. HPSv2 is used as the reward model.}
    \begin{tabular}{c|ccccc}
    \toprule
      Method & HPSv2 & PickScore & CLIPScore & VQAScore & ImageReward \\
    \midrule
    Base & 30.98 & 22.73 & 28.39 & 80.03 & 1.145 \\
    BoN & 32.75 & 23.06 & 28.86 & 81.77 & 1.252 \\
    DTO(Ours) & \textbf{41.02} & \textbf{24.16} & \textbf{28.99} & \textbf{83.09} & \textbf{1.279} \\
    \bottomrule
    \end{tabular}
    \label{tab:exp_flux}
\end{table}

DTO outperforms both the base model and BoN across all reported metrics, indicating that trajectory optimization remains effective on a current large-scale text-to-image model.

\subsection{Additional Results for SDE-Based Method}
\label{sec:app-sde-exp}

We implement SDE-based DTO on Stable-Diffusion-XL. The results are presented in Table \ref{tab:exp_sde}.

\begin{table}[htbp]
    \centering
    \caption{Results on HPDv2 test set. HPSv2 is used as reward model.}
    \begin{tabular}{c|ccccc}
    \toprule
      Method & HPSv2 & PickScore & CLIPScore & VQAScore & ImageReward \\
    \midrule
    Base & 28.19	& 22.58 & 30.78 & 82.60 & 0.874\\
    DTO(ODE) & 43.01 & 24.87 & 31.04 &  84.63 & \textbf{1.220} \\
    DTO(SDE) &  \textbf{43.03} & \textbf{24.89} & \textbf{31.09} & \textbf{84.84} & 1.204 \\
    \bottomrule
    \end{tabular}
    \label{tab:exp_sde}
\end{table}

As shown in the results, SDE-based DTO achieves comparable and occasionally superior performance compared to its ODE-based counterpart, demonstrating the flexibility and robustness of the proposed method.

\begin{figure}[htbp]
        \centering
        \begin{tcolorbox}[title=Human Annotation Criteria]

\textbf{Definition}:

Overall quality refers to the holistic assessment of a generated image based on both its visual quality (e.g., realism, clarity, aesthetics) and its alignment with the input text prompt (e.g., correctness, completeness, semantic consistency).

\textbf{Rating Scale (1–5 Likert Scale)}

5 – Excellent

The image is visually outstanding: sharp, clear, well-composed, and aesthetically pleasing.
Perfectly matches the text prompt in all key aspects (objects, attributes, relationships, style, and context).
No noticeable artifacts, distortions, or inconsistencies.
Demonstrates high realism or strong stylistic coherence (depending on prompt intent).

4 – Good

The image has high visual quality with only minor imperfections.
Largely consistent with the text prompt; most elements are correctly represented.
Minor omissions or slight inaccuracies may exist but do not significantly affect understanding.
Overall coherent and visually appealing.

3 – Fair

Acceptable visual quality but with noticeable issues (e.g., slight blur, minor artifacts, or composition weaknesses).
Partially aligned with the text prompt; some important elements are missing, incorrect, or ambiguous.
The image conveys the general idea but lacks precision or completeness.

2 – Poor

Low visual quality: visible artifacts, distortions, or poor composition.
Weak alignment with the text prompt; multiple key elements are missing, incorrect, or misrepresented.
The intended meaning of the prompt is only partially conveyed or confusing.

1 – Very Poor

Very low visual quality: severe artifacts, blurriness, or structural issues.
Fails to reflect the text prompt; major elements are absent or incorrect.
The image is largely unusable or irrelevant to the prompt.

\textbf{Evaluation Notes (for Annotators)}

Consider both dimensions jointly, not independently.

If visual quality is high but alignment is poor (or vice versa), assign a mid-range score (2–3) depending on severity.

Focus on key elements of the prompt rather than trivial details.

Avoid personal aesthetic bias unless explicitly required by the prompt.
        \end{tcolorbox}
\caption{Criteria of human annotation.}
\label{fig:criteria}
\end{figure}

\subsection{Details of Computational Cost}
\label{sec:app-latency}

We present the numerical results corresponding to Figure \ref{fig:latency} in Table \ref{tab:latency}.

\begin{table}[htbp]
    \centering
    \caption{Computation cost comparison of different methods. * denotes that the computation cost of BoN is calculated using batchified inference($N$ candidates are generated simultanously), resulting in less inference time yet higher GPU memory cost. SD1.5 is used as the base model.}
    \begin{tabular}{c|c|cc}
    \toprule
    Method & \makecell[c]{Inference Latency\\ (Seconds/Output)} ($\downarrow$) & GPU Memory ($\downarrow$) \\
    \midrule
    Base & 1.34 & 5G \\
    BoN & 6.30* & 20G* \\
    DNO & 427.38 & 18G \\
    DATE & 16.63  & 12G \\
    DTO(Ours) & 16.70 & 11G \\
    \bottomrule
    \end{tabular}
    
    \label{tab:latency}
\end{table}

The main comparison matches the number of reward evaluations, which is fixed to $10$ for BoN, DATE, and DTO. Under this protocol, batched BoN is faster in wall-clock time, while DTO achieves stronger alignment with lower GPU memory. Matching wall-clock time with DNO is not practical: DNO is about $10\times$ slower than DTO, so equalizing runtime would reduce DNO below its intended optimization budget.

To compare DTO with BoN under a similar wall-clock budget, we increase BoN to $N=20$ on SD1.5. The runtime is then approximately the same as DTO and DATE. Results on the HPDv2 test set are reported in Table~\ref{tab:bon20}. Even with this increased budget, BoN remains below DTO on all reported metrics, which is consistent with the limited scaling capability of BoN in Figure~\ref{fig:scale}.

\begin{table}[htbp]
    \centering
    \caption{Wall-clock-matched comparison of BoN and DTO on the HPDv2 test set. HPSv2 is used as the reward model. SD1.5 is used as the base model. $N=20$ brings the runtime of BoN close to DTO.}
    \begin{tabular}{c|ccccc}
    \toprule
      Method & HPSv2 & PickScore & CLIPScore & VQAScore & ImageReward \\
    \midrule
    BoN ($N=10$) & 27.48 & 21.35 & 30.21 & 76.92 & 0.565 \\
    BoN ($N=20$) & 28.14 & 21.45 & 30.23 & 78.01 & 0.643 \\
    DTO(Ours) & \textbf{40.75} & \textbf{23.41} & \textbf{30.51} & \textbf{79.20} & \textbf{0.702} \\
    \bottomrule
    \end{tabular}
    \label{tab:bon20}
\end{table}



\subsection{Details of Ablation Study}
\label{sec:app-abla}
The numeric results corresponding to Figure \ref{fig:abla} is presented in Table \ref{tab:abla}. 

\begin{table}[htbp]
    \centering
    \caption{Ablation results on HPDv2 test set. HPSv2 is used as reward model.}
    \begin{tabular}{c|ccccc}
    \toprule
       Method  & HPSv2 & PickScore & CLIPScore & VQAScore & ImageReward \\
    \midrule
    Base & 23.48 & 20.72 & 28.98 & 70.36 & 0.015 \\
    DTO(w/o TO) & 34.06 & 22.24 & 29.91 & 74.20 & 0.255 \\
    DTO(w/o ROSG) & \textbf{44.49} & \textbf{23.42} & 26.98 & 71.42 & -0.038 \\
    DTO(Ours) & 40.75 & 23.41 & \textbf{30.51} & \textbf{79.20} & \textbf{0.702}  \\
    \bottomrule
    \end{tabular}
    \label{tab:abla}
\end{table}

\begin{figure}
    \centering
    \includegraphics[width=0.6\linewidth]{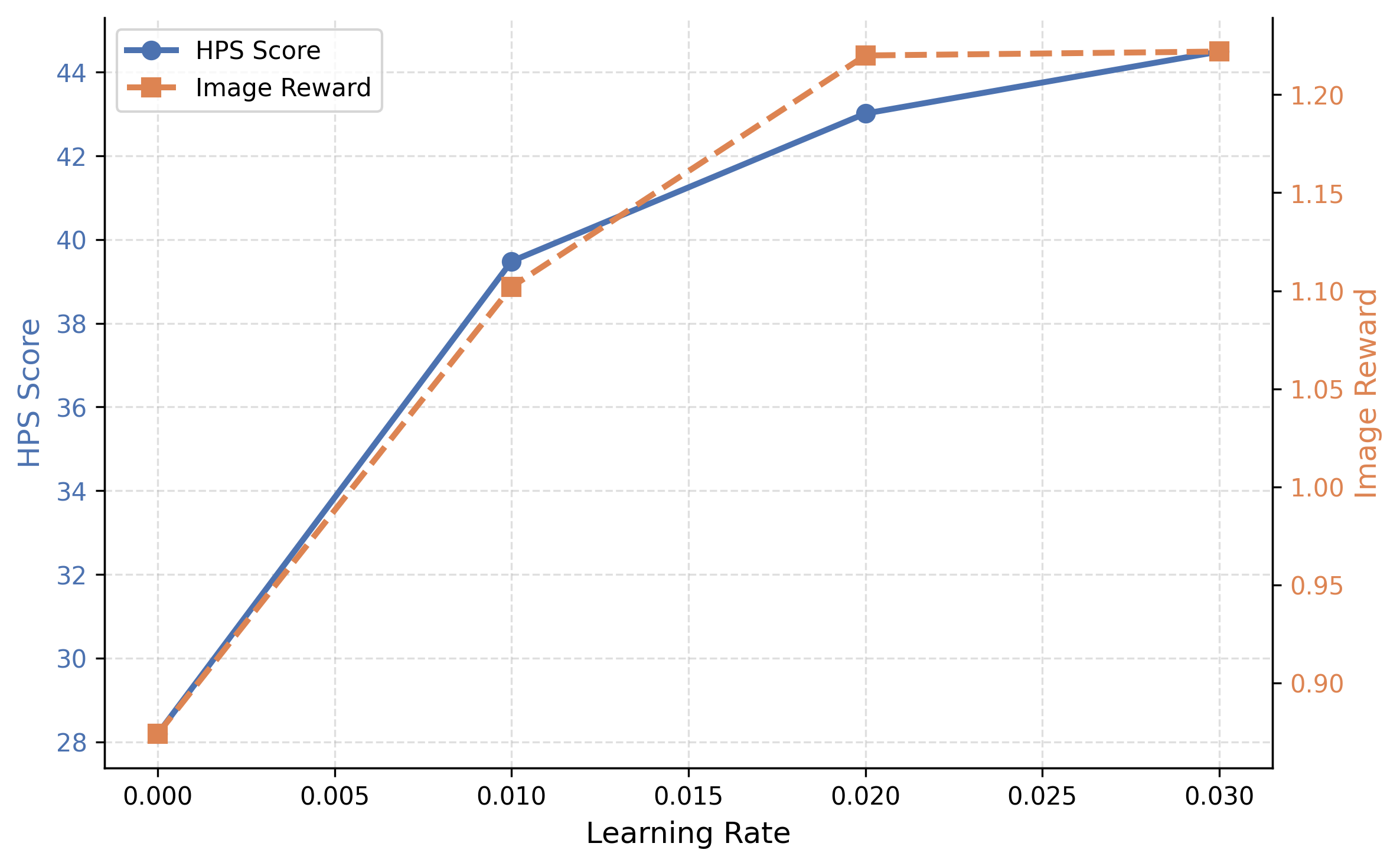}
    \caption{DTO performance under different learning rate selection.}
    \label{fig:lr}
\end{figure}

We further evaluate the impact of the learning rate. Experiments are conducted on SDXL, while keeping all other settings consistent with Section~\ref{sec:main_results}. The results are shown in Figure~\ref{fig:lr}. As observed, setting $\eta = 0.01, 0.02, 0.03$ leads to comparable performance, indicating that DTO is not overly sensitive to the choice of learning rate within a reasonable range.  Appropriate tuning of $\eta$ can further improve performance (e.g. setting $\eta = 0.03$ appears better than our default selection $\eta = 0.02$), which is an expected behaviour. Overall, these results demonstrate that DTO maintains stable performance as long as the learning rate is chosen within a reasonable range.

\paragraph{Distributional Drift of ROSG}
Rigorously certifying that ROSG stays on the pretrained distribution is difficult, because reward optimization intentionally allows some deviation from the pretrained sampler. We therefore evaluate whether the final generations remain realistic. FID is computed against MSCOCO-2014 training images, together with the relative score deviation $\frac{\|\epsilon_t - \epsilon'_t\|}{\|\epsilon'_t\|}$ (mean and max over $t$). Results on SD1.5 are reported in Table~\ref{tab:fid_drift}.

\begin{table}[htbp]
    \centering
    \caption{Score drift and FID on SD1.5. Drift is $\frac{\|\epsilon_t-\epsilon'_t\|}{\|\epsilon'_t\|}$, summarized by the mean and maximum over timesteps. FID is computed against MSCOCO-2014 training images. HPSv2 is used as the reward model.}
    \begin{tabular}{c|ccc}
    \toprule
      Method & Mean drift & Max drift & FID \\
    \midrule
    Base & -- & -- & 135.43 \\
    DTO (w/o ROSG) & 0.43 & 1.30 & 235.56 \\
    DTO (Full, ours) & 0.12 & 0.36 & \textbf{130.87} \\
    \bottomrule
    \end{tabular}
    \label{tab:fid_drift}
\end{table}

Without ROSG, FID increases substantially relative to the base model, and the generated image becomes noticeably noisier (Figure~\ref{fig:case_evolve}). This corresponds to the large-deviation regime in which the bound of Theorem~\ref{theo:combine} is no longer useful. With ROSG, the score deviation is reduced and FID returns to a level comparable to the base model, so optimization stays in the regime where the bound is more meaningful. If $x_t$ has already drifted far, the pretrained prediction $\epsilon'_t$ may itself be imperfect; ROSG is therefore best viewed as a practical stabilizer. The ablation and FID comparison still show that this stabilizer substantially improves final-sample quality.

\paragraph{Schedule of $w_t$}
The transition of $w_t$ around $t=T/2$ follows the same high-level design as adaptive guidance \citep{zhang2025much}. To validate this choice, we compare alternative schedules on SD1.5 under the same setting. \textit{Full Inter} uses $w_t=1-\frac{t}{T}$ for all timesteps, and \textit{Const} uses a constant $w_t=0.5$. Results are reported in Table~\ref{tab:w_schedule}. The proposed schedule in Equation~\ref{eq:scale_weight} performs best across all reported metrics.

\begin{table}[htbp]
    \centering
    \caption{Ablation of the ROSG weight schedule on the HPDv2 test set. HPSv2 is used as the reward model. SD1.5 is used as the base model.}
    \begin{tabular}{c|ccccc}
    \toprule
      Method & HPSv2 & PickScore & CLIPScore & VQAScore & ImageReward \\
    \midrule
    DTO (Full Inter) & 34.10 & 22.34 & 30.22 & 75.57 & 0.454 \\
    DTO (Const $w=0.5$) & 31.43 & 21.94 & 29.94 & 74.35 & 0.336 \\
    DTO (Ours) & \textbf{40.75} & \textbf{23.41} & \textbf{30.51} & \textbf{79.20} & \textbf{0.702} \\
    \bottomrule
    \end{tabular}
    \label{tab:w_schedule}
\end{table}

We also test a deviation-adaptive schedule. Let $\gamma_t=\frac{|\epsilon_t-\epsilon'_t|}{|\epsilon'_t|}$. We set $w_t=1$ if $\gamma_t<\gamma$ and $w_t=1-\frac{t}{T}$ if $\gamma_t\geq\gamma$, with $\gamma=0.1$ on SD1.5. Results are reported in Table~\ref{tab:w_gamma}. The threshold-based strategy is reasonable but lower than the original schedule. Selecting an optimal $\gamma$ is non-trivial \citep{zhang2025much}, so a similarity-threshold schedule can reduce practicality. The fixed step-based schedule is not necessarily optimal, but it transfers across different $T$ (SD1.5 with $T=50$, SD3 with $T=28$, and Stable-Diffusion-3.5-Large-Turbo with $T=4$) without per-model retuning.

\begin{table}[htbp]
    \centering
    \caption{Comparison of a score-deviation threshold schedule and the default $w_t$ on the HPDv2 test set. HPSv2 is used as the reward model. SD1.5 is used as the base model.}
    \begin{tabular}{c|ccccc}
    \toprule
      Method & HPSv2 & PickScore & CLIPScore & VQAScore & ImageReward \\
    \midrule
    DTO ($\gamma_t$ schedule) & 37.59 & 22.88 & 30.38 & 76.76 & 0.529 \\
    DTO (Ours) & \textbf{40.75} & \textbf{23.41} & \textbf{30.51} & \textbf{79.20} & \textbf{0.702} \\
    \bottomrule
    \end{tabular}
    \label{tab:w_gamma}
\end{table}

\paragraph{Which Part of the Trajectory to Optimize}
To isolate the benefit of full-trajectory optimization, we compare variants within the same DTO framework. ROSG is always applied, and all variants use the same number of reward evaluations. Results on SD1.5 are reported in Table~\ref{tab:traj_part}.

\begin{table}[htbp]
    \centering
    \caption{Ablation of the optimized trajectory variables on the HPDv2 test set. HPSv2 is used as the reward model. SD1.5 is used as the base model. ROSG is applied in every variant.}
    \begin{tabular}{c|ccccc}
    \toprule
      Method & HPSv2 & PickScore & CLIPScore & VQAScore & ImageReward \\
    \midrule
    DTO ($x_T$ only) & 37.49 & 22.82 & 30.33 & 76.90 & 0.508 \\
    DTO (First Half) & 36.65 & 22.68 & 30.26 & 76.72 & 0.473 \\
    DTO (Second Half) & 24.25 & 20.81 & 29.35 & 71.74 & 0.077 \\
    DTO (Full, Ours) & \textbf{40.75} & \textbf{23.41} & \textbf{30.51} & \textbf{79.20} & \textbf{0.702} \\
    \bottomrule
    \end{tabular}
    \label{tab:traj_part}
\end{table}

Full-trajectory DTO achieves the best overall performance. Optimizing $x_T$ alone or only the first half already yields gains, whereas optimizing only the second half performs poorly. Early denoising steps are therefore more important than late steps, which is consistent with the proposed $w_t$ schedule. Inference time stays similar across these variants. The main bottleneck is ROSG, which requires sequential computation over timesteps, so restricting the optimized variables to $x_T$, the first half, or the second half does not substantially reduce latency. A more practical way to reduce inference time is to reduce the number of reward evaluations, which still yields strong results as shown in Figure~\ref{fig:scale}.

\subsection{Details of Human Evaluation}
\label{sec:app-human_eval}
We recruit two well-educated annotators to evaluate 50 randomly selected samples from the generation results on the HPDv2 test set. The annotators are informed of how their annotations will be used, participate on a voluntary basis, and are fairly compensated. Inter-annotator agreement is measured using Krippendorff’s alpha \citep{krippendorff2011computing}, yielding a score of 0.768, which indicates substantial agreement. Numerical results corresponding to Figure \ref{fig:human_eval} is provided in Table \ref{tab:exp_human_eval}. The detailed annotation guidelines are provided in Figure~\ref{fig:criteria}.

\begin{table}[htbp]
    \centering
    \caption{Results of human evaluation.}
    \begin{tabular}{cccccc}
    \toprule
        Base & BoN & ReNeg & DATE & DNO & DTO(Ours) \\
    \midrule
        3.00 & 3.31 & 3.35 & 3.36 & 3.33 & \textbf{3.60} \\
    \bottomrule
    \end{tabular}
    \label{tab:exp_human_eval}
\end{table}

We additionally conduct a blind human evaluation on another 50 samples generated with SD3, following the same protocol. Only the base model, BoN, and DTO are compared, matching the methods evaluated on SD3 in the main experiments. Results are reported in Table~\ref{tab:exp_human_eval_sd3}. DTO remains preferred over the base model and BoN.

\begin{table}[htbp]
    \centering
    \caption{Human evaluation on 50 SD3 samples from the HPDv2 test set, using the same protocol as Table~\ref{tab:exp_human_eval}.}
    \begin{tabular}{ccc}
    \toprule
        Base & BoN & DTO(Ours) \\
    \midrule
        4.04 & 4.14 & \textbf{4.26} \\
    \bottomrule
    \end{tabular}
    \label{tab:exp_human_eval_sd3}
\end{table}

\subsection{More Case Studies}
\label{sec:app-case}
\begin{figure}
    \centering
    \begin{subfigure}[b]{0.13\textwidth}
        \centering
        \includegraphics[width=1.0\linewidth]{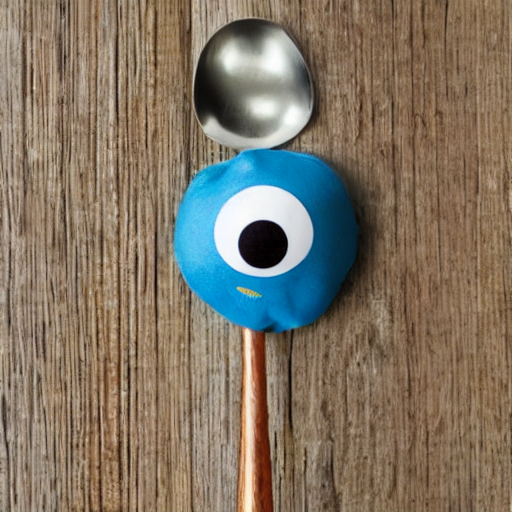}
        \caption{Base \\ ~}
    \end{subfigure}
    \hfill
     \begin{subfigure}[b]{0.13\textwidth}
        \centering
        \includegraphics[width=1.0\linewidth]{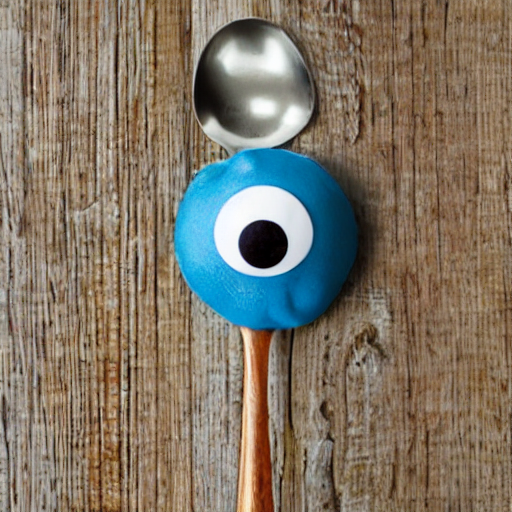}
        \caption{DTO \\ (2 steps)}
    \end{subfigure}
    \hfill
    \begin{subfigure}[b]{0.13\textwidth}
        \centering
        \includegraphics[width=1.0\linewidth]{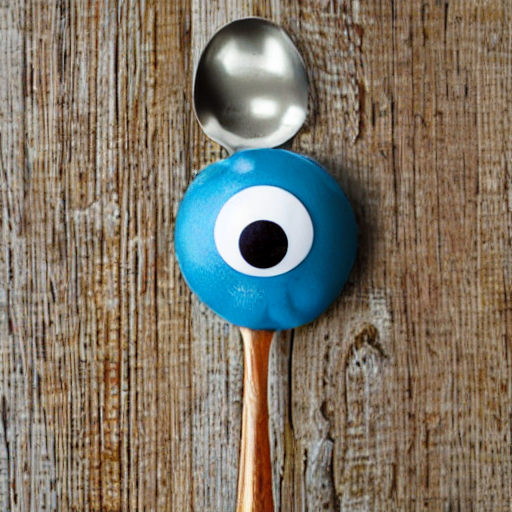}
        \caption{DTO \\ (4 steps)}
    \end{subfigure}
    \hfill
     \begin{subfigure}[b]{0.13\textwidth}
        \centering
        \includegraphics[width=1.0\linewidth]{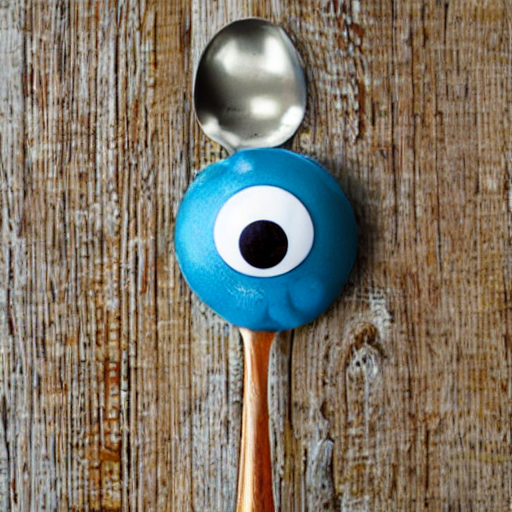}
        \caption{DTO \\ (6 steps)}
    \end{subfigure}
    \hfill
    \begin{subfigure}[b]{0.13\textwidth}
        \centering
        \includegraphics[width=1.0\linewidth]{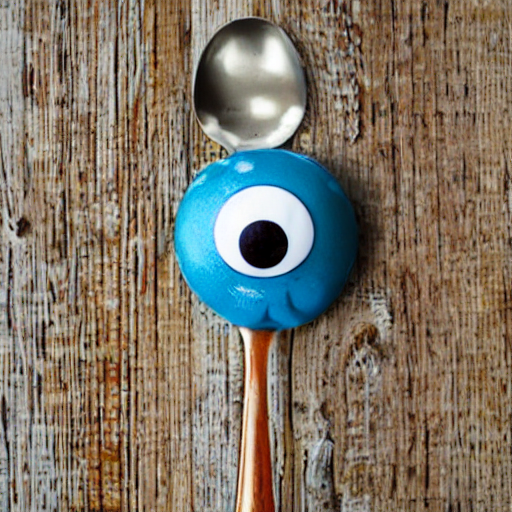}
        \caption{DTO \\ (8 steps)}
    \end{subfigure}
    \hfill
     \begin{subfigure}[b]{0.13\textwidth}
        \centering
        \includegraphics[width=1.0\linewidth]{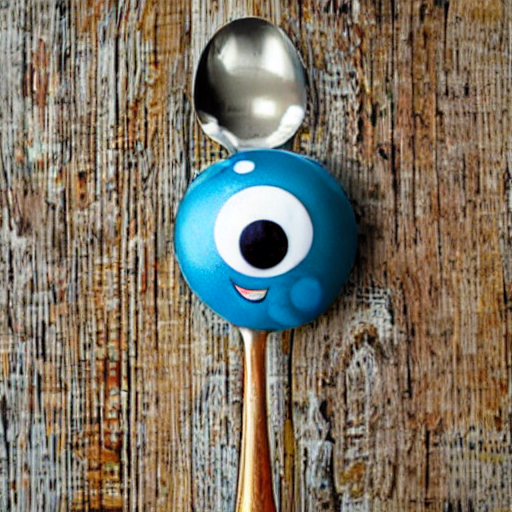}
        \caption{DTO \\ (10 steps)}
    \end{subfigure}
    \hfill
    \begin{subfigure}[b]{0.13\textwidth}
        \centering
        \includegraphics[width=1.0\linewidth]{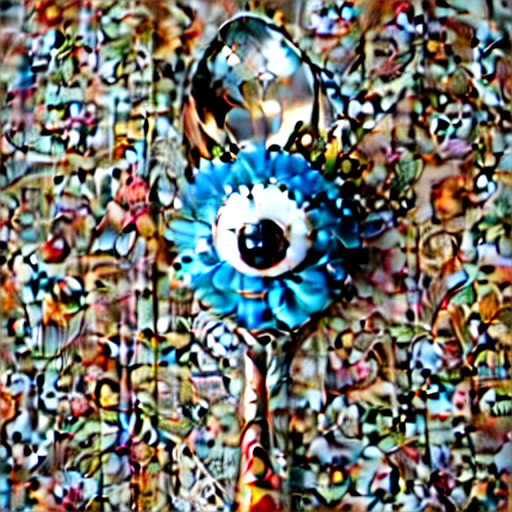}
        \caption{DTO \\ (w/o ROSG)}
    \end{subfigure}
    \caption{Generation results of DTO under different setups based on SD1.5. The prompt is ``A spoon dressed up with eyes and a smile.''}
     \label{fig:case_evolve}
\end{figure}

We present more case studies in this part. Firstly, we show how the generation result evolves as DTO progresses in Figure \ref{fig:case_evolve}. As can be seen, as the number of optimization steps increases, there are two major changes. Firstly, the texture of the background wooden surface is better rendered. Secondly, the lower-left of the eye in the image is becoming a clearer smiling mouth, better aligning with the textual prompt. Both changes suggest the effectiveness of DTO. We also present the DTO without ROSG in the last image, and as can be seen, the generated result is noisy, indicating an OOD problem, further suggesting the importance of introducing ROSG.

\begin{figure}
    \centering
    \includegraphics[width=0.5\linewidth]{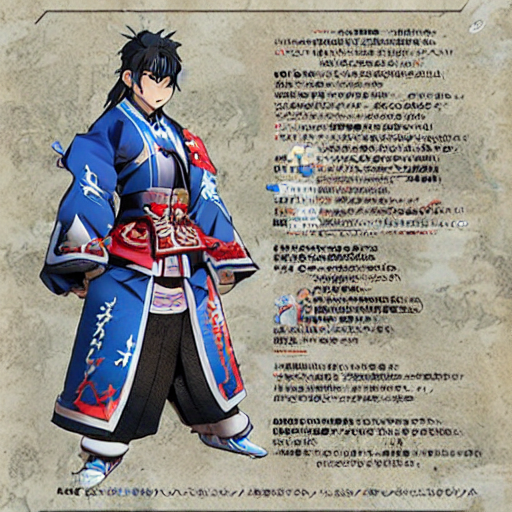}
    \caption{A failure case generated by DTO using SD1.5 as the base model. The prompt is ``Keqing from Genshin Impact.''}
    \label{fig:fail_case}
\end{figure}

\paragraph{Failure Case Analysis}
We additionally present a failure case where DTO fail to improve the overall generation quality in Figure \ref{fig:fail_case}. As can be seen from the image, the character in the image has nothing to do with Keqing from Genshin Impact. We attribute this failure to the poor initialization of the trajectory to optimize. Since the pretrained diffusion model has little knowledge about this character, the trajectory initialized using the pretrained diffusion model is still hard to optimize, resulting in an unsatisfactory generation result. This is also a common failure in most inference-time optimization methods \citep{zhang-wan-2025-r}.

\end{document}